\documentclass[10pt,twocolumn,letterpaper]{article}

\usepackage{cvpr}              

\definecolor{cvprblue}{rgb}{0.21,0.49,0.74}
\usepackage[pagebackref,breaklinks,colorlinks,allcolors=cvprblue]{hyperref}

\def\paperID{3294} 
\def\confName{CVPR}
\def\confYear{2025}

\usepackage{hyperref}
\usepackage{url}            
\usepackage{booktabs}       
\usepackage{amsfonts}       
\usepackage{nicefrac}       
\usepackage{microtype}      

\usepackage{multirow}
\usepackage{graphicx}
\usepackage{amsmath}
\usepackage{amssymb}
\usepackage{mathtools, nccmath}
\usepackage{amsthm}
\usepackage{pifont}

\usepackage{algorithm, algorithmic}
\usepackage{graphicx}
\usepackage{wrapfig, lipsum, booktabs}
\usepackage{colortbl}
\usepackage{comment}
\usepackage{marvosym}
\usepackage{enumitem}
\usepackage{tabularx}
\usepackage{nicematrix}
\usepackage{caption}
\usepackage{diagbox}
\usepackage{dsfont}
\usepackage{appendix}

\def \flame {\raisebox{-.1\height}{\includegraphics[height=0.7\baselineskip]{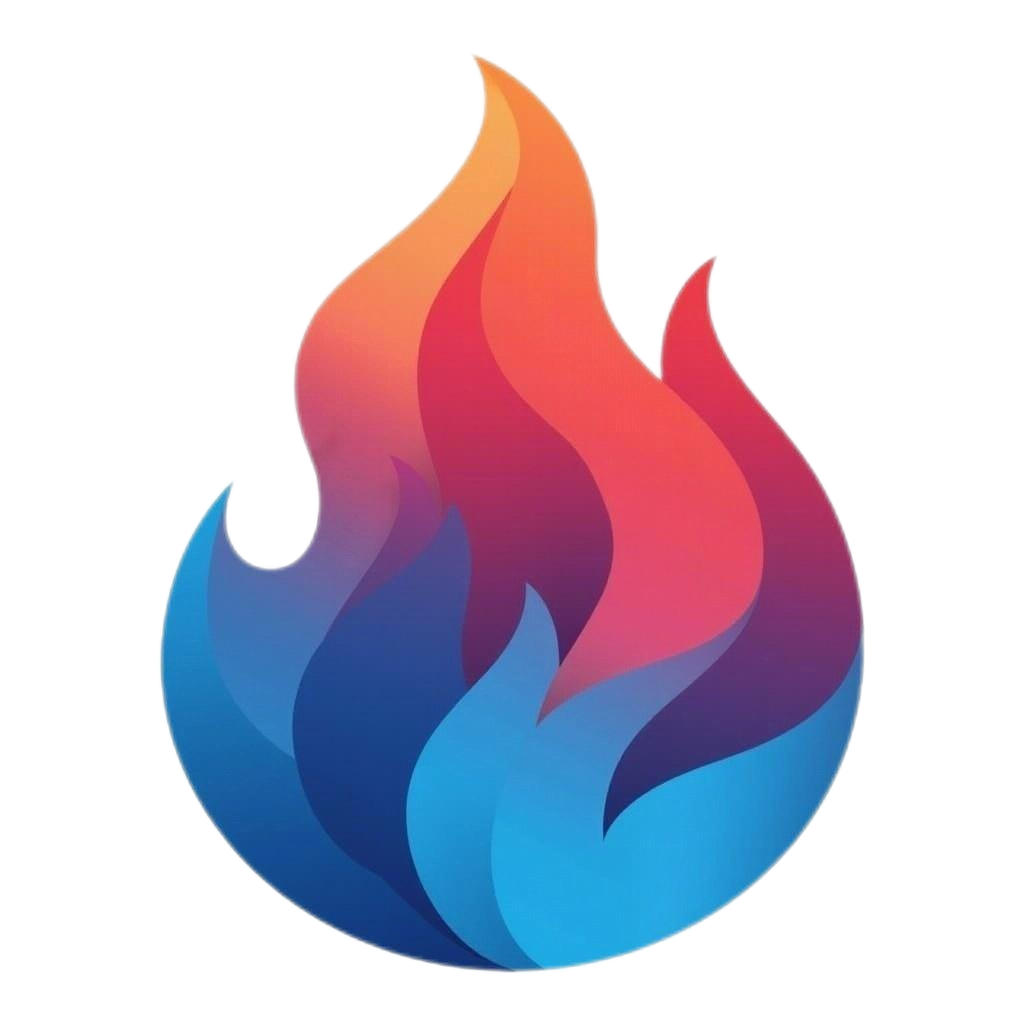}}\xspace}

\usepackage[dvipsnames]{xcolor}
\colorlet{darkgreen}{green!65!black}
\colorlet{darkblue}{blue!75!black}
\colorlet{darkred}{red!80!black}
\definecolor{lightblue}{HTML}{0071bc}
\definecolor{lightgreen}{HTML}{39b54a}
\definecolor{purple1}{HTML}{b71a3b}
\definecolor{blue1}{HTML}{95bddc}

\usepackage[accsupp]{axessibility}

\title{FLAME\flame: Frozen Large Language Models Enable Data-Efficient Language-Image Pre-training}

\author{Anjia Cao\textsuperscript{1} \qquad
Xing Wei\textsuperscript{1} \qquad
Zhiheng Ma\textsuperscript{2,3,4}\thanks{Corresponding author: Zhiheng Ma.}\\[2mm]
{\small \textsuperscript{1}School of Software Engineering, Xi'an Jiaotong University \qquad}
{\small \textsuperscript{2}Shenzhen University of Advanced Technology \qquad}\\
{\small \textsuperscript{3}Guangdong Provincial Key Laboratory of Computility Microelectronics \qquad}\\
{\small \textsuperscript{4}Shenzhen Institute of Advanced Technology, Chinese Academy of Sciences\qquad}\\
{\tt\small caoanjia7@stu.xjtu.edu.cn\qquad weixing@mail.xjtu.edu.cn\qquad mazhiheng@suat-sz.edu.cn}
}

\begin{document}
\maketitle

\begin{abstract}

Language-image pre-training faces significant challenges due to limited data in specific formats and the constrained capacities of text encoders. While prevailing methods attempt to address these issues through data augmentation and architecture modifications, they continue to struggle with processing long-form text inputs, and the inherent limitations of traditional CLIP text encoders lead to suboptimal downstream generalization. In this paper, we propose FLAME (Frozen Large lAnguage Models Enable data-efficient language-image pre-training) that leverages frozen large language models as text encoders, naturally processing long text inputs and demonstrating impressive multilingual generalization. FLAME comprises two key components: 1) a multifaceted prompt distillation technique for extracting diverse semantic representations from long captions, which better aligns with the multifaceted nature of images, and 2) a facet-decoupled attention mechanism, complemented by an offline embedding strategy, to ensure efficient computation. Extensive empirical evaluations demonstrate FLAME's superior performance. When trained on CC3M, FLAME surpasses the previous state-of-the-art by 4.9\% in ImageNet top-1 accuracy. On YFCC15M, FLAME surpasses the WIT-400M-trained CLIP by 44.4\% in average image-to-text recall@1 across 36 languages, and by 34.6\% in text-to-image recall@1 for long-context retrieval on Urban-1k. Code is available at \url{https://github.com/MIV-XJTU/FLAME}.

\end{abstract}
\section{Introduction}
\begin{figure}[!t]
\centering
\includegraphics[width=\linewidth]{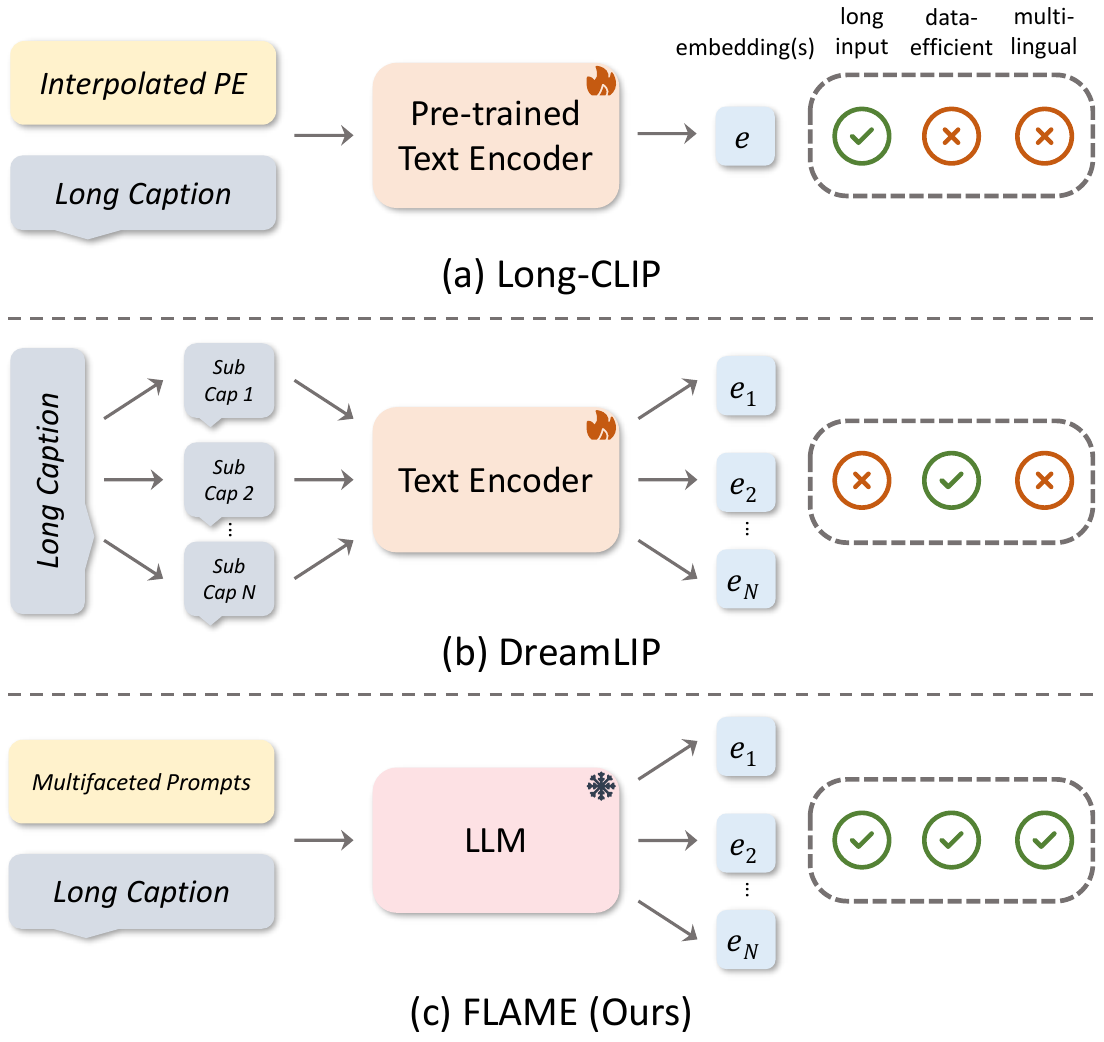}
    \vspace{-12pt}
    \caption{\textbf{Conceptual comparison of text streams.} FLAME leverages frozen large language models (LLMs) to directly process long captions. With multifaceted prompts, this framework extracts diverse semantic embeddings, achieving data efficiency. Preserving LLMs' inherent capabilities enables multilingual generalization.}
    \label{fig:comparison}
    \vspace{-1em}
\end{figure}

Multimodal learning, particularly language-image pre-training, has made significant strides with models like CLIP (Contrastive Language-Image Pre-training) \cite{radford2021clip}, which learn transferable visual representations through language supervision. CLIP-style models have demonstrated state-of-the-art performance across a wide range of downstream tasks \cite{radford2021clip, zhai2023siglip, li2023clipa, sun2023eva-clip, li2022glip, xu2022groupvit, rombach2022stablediffusion}. However, the deployment of CLIP models in real-world scenarios presents several challenges, especially when data is scarce. Specifically, there are two major limitations that hinder their broader applicability: 1) \textbf{Limited Training Pairs}—high-quality image-text pairs, particularly those involving long-form descriptions and non-English languages, are difficult to come by; and 2) \textbf{Constrained Model Capacity}—the standard CLIP text encoder has a max sequence length limitation of 77 tokens, which restricts its ability to leverage the rich, long-context data that could be highly beneficial.

To address these challenges, recent works have primarily focused on data-centric solutions. Approaches such as synthesizing high-quality short captions using multimodal large language models \cite{fan2024laclip, lai2023veclip, liu2023mllm-a} or translating short captions into multiple languages \cite{visheratin2023nllb-clip, yang2022cn-clip, carlsson2022mclip} attempt to augment training data. Some works also attempt to break down long captions into shorter segments \cite{zheng2024dreamlip} to fit the constraints of the CLIP text encoder. While these solutions are promising, they suffer from inherent limitations: The capacity bottleneck of the text encoder, particularly strict input restrictions, constrains the efficient utilization of data, and the naive decomposition strategy may lead to a loss of contextual meaning, resulting in suboptimal representations.

On the other hand, model-centric approaches—such as interpolating positional encodings to handle longer sequences \cite{zhang2024long-clip}—attempt to extend the capabilities of CLIP's text encoder. However, these solutions are fundamentally constrained by the design of the CLIP model, which allows a maximum of 77 tokens, with an effective token length of fewer than 20. These limitations often translate into suboptimal performance on downstream tasks \cite{koukounas2024jinaclip}. Consequently, there remains a need for a more effective text encoder capable of processing long-form and multilingual content without relying on extensive data augmentation or architectural modifications.

In this paper, we propose a fundamentally new approach by utilizing \textbf{frozen large language models (LLMs)} as text encoders. Our key insight is that despite the conventional wisdom suggesting frozen text encoders lead to suboptimal performance~\cite{zhai2022lit, fan2024laclip}, powerful LLMs can provide sufficiently rich semantic representations for effective language-image pre-training. 
This strategy holds significant promise due to the inherent capabilities of LLMs: they are pre-trained on diverse multilingual corpora, excel at processing long-form text, and effectively extract semantic information at various granularities. While traditional decoder-only LLMs use causal attention mechanisms that limit their ability to generate rich contextual representations, recent advancements have shown that carefully designed prompting and attention mechanisms can enable effective text embedding for various tasks \cite{jiang2023prompteol, zhuang2024promptreps, zhang2024pretcot, springer2024echo, behnamghader2024llm2vec}. However, applying these techniques to language-image pre-training introduces unique challenges. Unlike general text embedding tasks where a single representation may suffice, visual content is inherently more complex, containing multifaceted semantic information that requires comprehensive understanding, especially when training data is limited.

To address these challenges, we introduce \textbf{FLAME} (Frozen Large lAnguage Models Enable data-efficient language-image pre-training), an innovative framework that leverages frozen LLMs to enhance language-image pre-training efficiency. FLAME employs a novel \textbf{multifaceted prompt distillation} technique to guide frozen LLMs in extracting various semantic facets from each image-text pair, maximizing the utility of limited training data. For example, when processing a long caption, FLAME simultaneously extracts semantic features ranging from fine-grained object attributes to abstract scene-level concepts. This approach facilitates a comprehensive visual-semantic alignment for each training sample, effectively enabling long-context understanding.

To ensure computational efficiency, FLAME incorporates a \textbf{facet-decoupled attention mechanism} that enables single-pass inference, preserving the independence of different semantic facets. This attention mechanism, combined with an offline embedding strategy, minimizes training overhead and improves the practical viability of our framework. By keeping the LLMs frozen, FLAME maintains the long-context and multilingual processing capabilities inherent in LLMs while reducing the need for expensive fine-tuning.

We evaluate FLAME across various data-scarce scenarios, demonstrating its superiority over previous methods. When trained on the CC3M dataset, FLAME outperforms the state-of-the-art model \cite{zheng2024dreamlip} by achieving a 4.9\% improvement in ImageNet top-1 accuracy. In multilingual evaluation, FLAME surpasses WIT-400M-trained CLIP \cite{radford2021clip} by 44.4\% in average image-to-text recall@1 across 36 languages on the Crossmodal-3600 dataset \cite{thapliyal2022crossmodal-3600}. Moreover, FLAME excels in long-context image-text retrieval tasks, achieving a text-to-image recall@1 of 87.9\% on the Urban-1k dataset \cite{zhang2024long-clip}, outperforming WIT-400M-trained CLIP by 34.6\%. These results are particularly impressive considering that FLAME achieves these outcomes with a fraction of the training data used by previous methods.

Our key contributions are as follows:
\begin{itemize}
    \item We challenge the conventional wisdom about frozen text encoders and demonstrate that frozen LLMs can effectively enhance language-image pre-training through rich semantic representations.
    \item \textbf{FLAME} introduces a novel framework that leverages frozen LLMs for data-efficient language-image pre-training via \textbf{multifaceted prompt distillation}.
    \item We propose a \textbf{facet-decoupled attention mechanism}, complemented by an offline embedding strategy, to enhance computational efficiency while capturing comprehensive semantic representations, making the framework practical for real-world applications.
    \item We demonstrate through extensive experimental validation that FLAME significantly outperforms existing methods in data-scarce scenarios, excelling in long-context understanding and multilingual tasks.
\end{itemize}

\section{Related Works}
\paragraph{Contrastive Language-Image Pre-training.}
CLIP~\cite{radford2021clip} pioneered transferable visual representation learning through language supervision, with subsequent works like ALIGN~\cite{jia2021align} scaling up through larger datasets. Recent efforts have focused on improving CLIP's efficacy and data efficiency through various techniques: self-supervision~\cite{mu2022slip,li2021declip}, modified losses~\cite{yao2021filip,gao2022pyramidclip,yu2022coca,lee2022uniclip,kim2023mcd}, selective encoder tuning~\cite{zhai2022lit}, and masked training~\cite{li2023flip,dong2023maskclip}. To enhance representation quality, researchers have explored hierarchical semantics~\cite{geng2022hiclip}, relaxed matching constraints~\cite{gao2024softclip}, and caption synthesis using multimodal models~\cite{yang2023alip, fan2024laclip,tian2023stablerep,lai2023veclip,liu2023mllm-a,zheng2024dreamlip}. Efforts to handle long-form text~\cite{zhang2024long-clip} and multilingual content~\cite{yang2022cn-clip,visheratin2023nllb-clip,carlsson2022mclip,chen2023altclip} have relied on position encoding modifications and extensive data synthesis. In contrast to these approaches that work within CLIP's inherent constraints, our framework fundamentally resolves both the length limitation and multilingual challenges by leveraging the rich capabilities of frozen LLMs.

\paragraph{Large Language Models as Text Encoders.}
While large language models (LLMs)~\cite{brown2020gpt3,achiam2023gpt4,touvron2023llama2,bai2023qwen,jiang2023mistral,abdin2024phi3,bai2024atlas} excel at text understanding, their decoder-only architecture with causal attention poses challenges for generating rich text representations. Recent works have explored various ways to extract embeddings from LLMs, including contrastive training on paired data~\cite{wang2023e5mistralinst,lee2024gecko, muennighoff2024grit}, single-word distillation~\cite{jiang2023prompteol,zhuang2024promptreps,lei2024metaeol,zhang2024pretcot}, sentence repetition~\cite{springer2024echo}, and bidirectional attention with modified training objectives~\cite{behnamghader2024llm2vec}. Our work advances this direction by introducing multifaceted prompt distillation and facet-decoupled attention, specifically designed for language-image pre-training. Unlike previous methods that focus on general text embedding, our approach extracts comprehensive visual-semantic representations spanning multiple granularities while maintaining computational efficiency through innovative attention mechanisms.

\section{Method}
\label{sec:method}

Language-image pre-training faces two fundamental challenges: limited availability of high-quality training pairs (especially for long-context and multilingual scenarios) and restricted token length of text encoders. While previous works address these challenges through data synthesis~\cite{fan2024laclip, liu2023mllm-a, lai2023veclip, zheng2024dreamlip} or architecture modifications~\cite{zhang2024long-clip}, these solutions heavily remain constrained by CLIP's fundamental design limitations.

We propose FLAME (Frozen Large lAnguage Models Enable data-efficient language-image pre-training), introducing a paradigm shift by leveraging frozen large language models (LLMs) as text encoders. This approach challenges the conventional wisdom that frozen text encoders lead to suboptimal performance~\cite{zhai2022lit, fan2024laclip}. We demonstrate that powerful frozen LLMs, when properly utilized, can enable effective language-image pre-training through their rich semantic representations. FLAME inherits LLMs' native capabilities in processing long-form text and understanding multiple languages, while enabling efficient training through optimized text embedding computation. 

\subsection{Framework Overview}
\label{subsec:training}
\begin{figure}[!t]
\centering
\includegraphics[width=\linewidth]{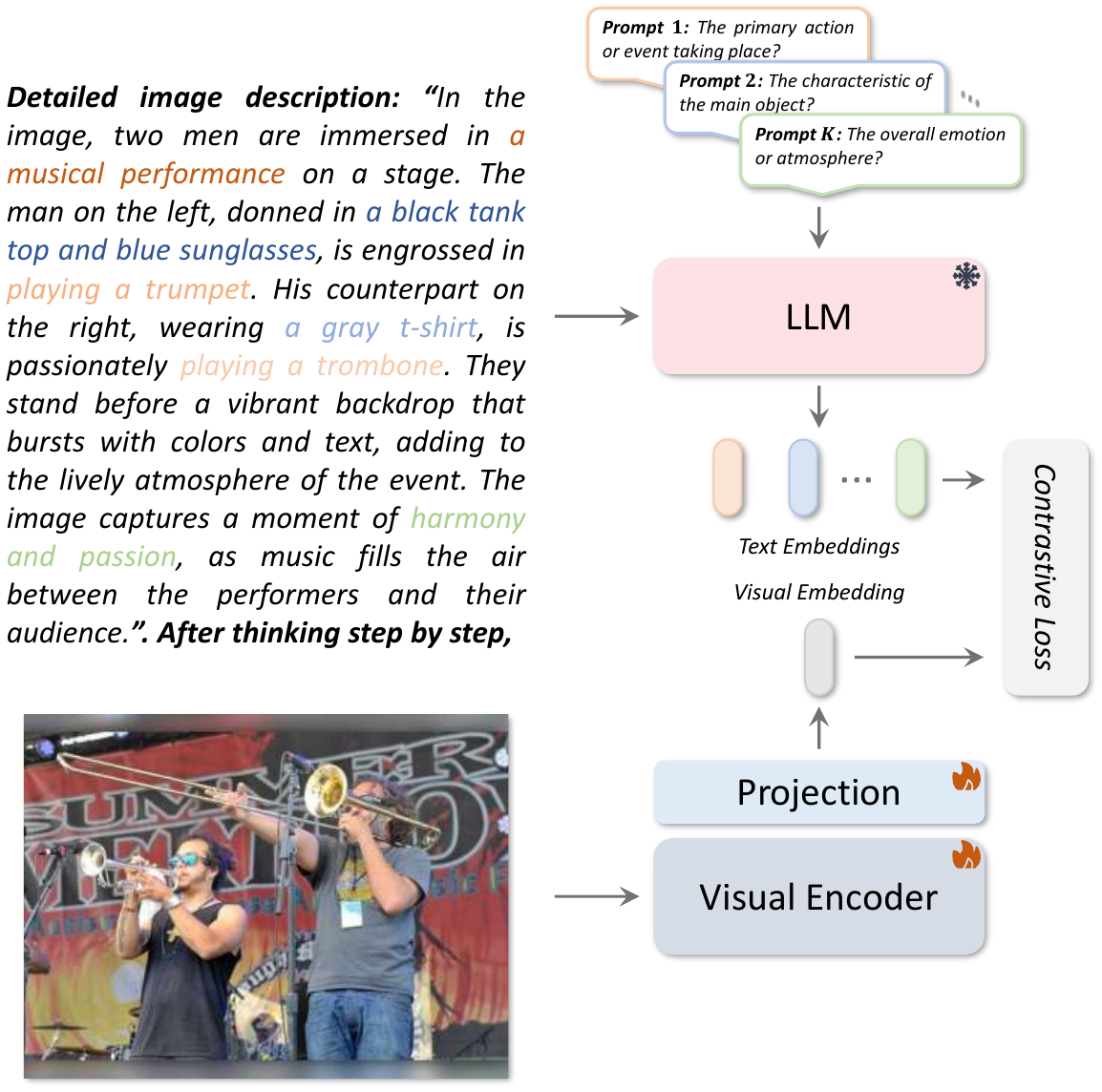}
    \vspace{-12pt}
    \caption{\textbf{FLAME overview.} This framework harnesses the sophisticated long-text comprehension capabilities of large language models to conduct language-image pre-training directly on long captions. Based on multifaceted prompts, it extracts a diverse array of representations embedded within the long caption, thereby enhancing semantic alignment.}
    \label{fig:overview}
    \vspace{-1em}
\end{figure}

FLAME leverages frozen LLMs for visual-semantic alignment by exploiting their sophisticated text comprehension capabilities without any fine-tuning. As illustrated in Figure~\ref{fig:overview}, our framework consists of three key components: 1) a frozen LLM that transforms each long caption into \textbf{multiple} complementary semantic representations through our proposed multifaceted prompt distillation technique, 2) a trainable visual encoder that distills the image content into a unified global representation, and 3) a learnable projection $\phi$ implemented as a two-layer multilayer perceptron (MLP) that aligns these heterogeneous embeddings in a shared semantic space for contrastive learning.

Given a batch of $N$ image-text pairs $\{(x_i,y_i)\}_{i=1}^N$, where $x_i$ represents an image and $y_i$ is its corresponding long caption, we extract multiple semantic perspectives using a set of prompts $\{P_k\}_{k=1}^K$. The training objective is a symmetric contrastive loss: $\mathcal{L}=\frac{1}{2}(\mathcal{L}_{I} + \mathcal{L}_{T})$, where the text-to-image component $\mathcal{L}_I$ is formulated as:
\begin{equation}
    \small
    \mathcal{L}_I = -\sum_{i=1}^{N}\sum_{k=1}^{K} \log \frac{\exp(\cos\langle f_t(y_i, P_k), \phi(f_v(x_i)) \rangle / \tau)}{\sum_{j=1}^{N}\exp(\cos\langle f_t(y_j, P_k), \phi(f_v(x_i)) \rangle / \tau)},
\end{equation}
where $f_t(y_i,P_k) \in \mathbb{R}^{d_t}$ generates text embeddings through the LLM with prompt $P_k$, $f_v(x_i) \in \mathbb{R}^{d_v}$ extracts visual features through the encoder, $\phi(\cdot)$ is the MLP that projects visual features into the text embedding space, and $\tau$ is the learnable temperature. The image-to-text loss $\mathcal{L}_T$ follows the same formulation with image and text roles reversed.

The following sections detail our two key technical contributions: multifaceted prompt distillation for extracting diverse semantic representations from long captions (Section~\ref{subsec:mpd}), and facet-decoupled attention for efficient computation (Section~\ref{subsec:fda}).

\subsection{Multifaceted Prompt Distillation}
\label{subsec:mpd}

While LLMs have shown promising capabilities in natural language understanding, their decoder-only architecture with causal attention inherently limits their ability to generate rich contextual representations. Previous works attempt to address this limitation either through bidirectional attention retraining~\cite{behnamghader2024llm2vec} or by extracting the final token's features~\cite{jiang2023prompteol, zhuang2024promptreps, zhang2024pretcot}. However, these approaches only produce a single universal representation, which is insufficient for language-image pre-training where images contain multiple semantic facets from objects to scenes. To address this challenge, we introduce multifaceted prompt distillation, a systematic approach that decomposes visual semantics into a hierarchical structure mirroring human visual perception, from local object properties to global scene understanding. This semantic decomposation aligns with well-established visual understanding paradigms~\cite{li2017scene, li2024scene, onoe2024docci}.

\paragraph{Hierarchical Semantic Decomposition.} We identify three fundamental levels of visual semantics that are crucial for comprehensive image understanding:

\begin{itemize}
    \item \textit{Entity Level:} Captures both primary and secondary objects in the scene, essential for object-centric understanding. This level is further divided into object categories and their distinctive attributes, enabling fine-grained visual discrimination.
    
    \item \textit{Interaction Level:} Focuses on dynamic elements such as actions and events, crucial for understanding the temporal and relational aspects of the scene. This bridges the gap between static object recognition and dynamic scene interpretation.
    
    \item \textit{Scene Level:} Abstracts high-level concepts such as atmosphere and emotion, capturing the holistic context humans naturally perceive beyond individual objects and actions.
\end{itemize}

\begin{figure}[!t]
\centering
\includegraphics[width=\linewidth]{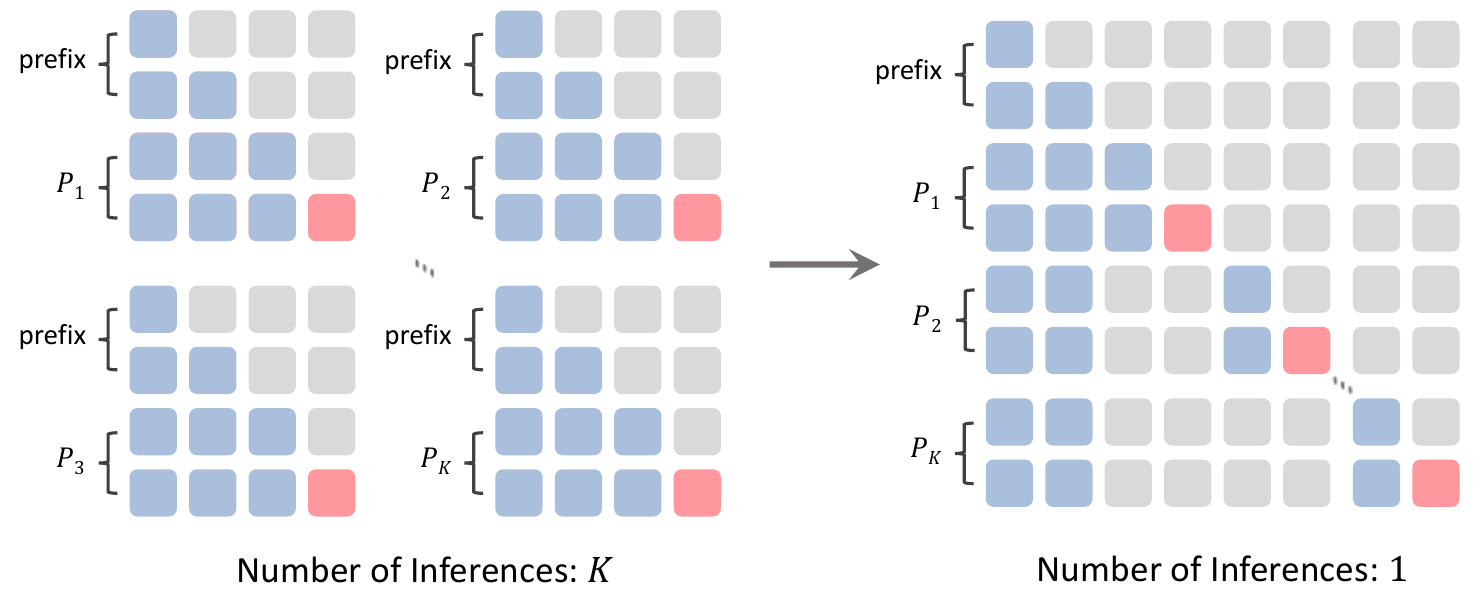}
    \vspace{-18pt}
    \caption{\textbf{Facet-decoupled attention.} By streamlining all prompts with a shared prefix and applying this facet-decoupled attention mask, the overhead of feature extraction is greatly reduced. The positions in \textcolor{darkred}{red} indicate the features to be extracted.}
    \label{fig:attention_mask}
    \vspace{-1.5em}
\end{figure}

\paragraph{Prompt Engineering Principles.} Based on this hierarchical decomposition, we design our prompts following three key principles:

\begin{itemize}
    \item \textit{Semantic distinctiveness:} Each prompt targets a distinct aspect of visual semantics to minimize redundancy in the extracted representations.
    
    \item \textit{Constrained Output Space:} All prompts are designed to generate single-word responses, forcing the LLM to distill complex visual concepts into concise, discriminative features.
    
    \item \textit{Cognitive Alignment:} Prompt structure mirrors human visual processing by explicitly separating different levels of semantic abstraction.
\end{itemize}

Our prompt set $\{P_k\}_{k=1}^{K}$ systematically covers the identified semantic hierarchy. Each prompt follows a template structure as shown below, where text in \textcolor{lightblue}{blue} indicates the shared prefix containing the image caption $y_i$, and \textcolor{darkred}{red} marks the position where we extract the final hidden state feature from LLM:

\begin{center}
\texttt{\textcolor{lightblue}{Detailed image description: "$y_i$". After thinking step by step,} the prominent characteristic or pattern of the main object in this image means in just one word:\textcolor{darkred}{"}}
\end{center}

This hierarchical decomposition enables more robust visual-semantic alignment by matching images and text at multiple semantic levels simultaneously, while providing interpretable insights into the model's understanding of visual content through its structured representation of semantic concepts. The complete list of our prompts is provided in the supplementary material.

\subsection{Facet-Decoupled Attention}
\label{subsec:fda}
While our multifaceted approach provides rich semantic representations, a naive implementation would require $K$ separate forward passes through the LLM, resulting in significant computational overhead. To achieve efficient multi-perspective feature extraction, we propose an innovative attention mechanism design that enables single-pass inference while preserving semantic independence.

Our key insight is that all prompts share an identical prefix structure which dominates the input sequence length. Specifically, each prompt contains the core context: ``Detailed image description: $y_i$. After thinking step by step,". This observation motivates us to transform $K$ independent sequences into a single structured input:
$$\{\langle y_i, \mathrm{prefix}, P_k \rangle\}_{k=1}^K \rightarrow \langle y_i, \mathrm{prefix}, \{P_k \}_{k=1}^K \rangle$$

However, with this concatenated structure, standard causal attention would allow information leakage between different semantic facets, compromising the independence of extracted features. We address this challenge through facet-decoupled attention, which enforces strict boundaries between semantic perspectives using a specially designed attention mask (Figure~\ref{fig:attention_mask}). The mask is implemented through an efficient prefix KV Cache distribution strategy: by computing the prefix's KV Cache once and distributing it for parallel prompt computation, our mechanism enables efficient extraction of multifaceted features while maintaining their semantic distinctiveness.

Notably, our use of a frozen LLM enables a significant practical advantage: we can pre-compute and cache all prompt-specific text embeddings offline. This design choice eliminates the LLM forward passes during training, resulting in competitive memory consumption and training speed compared to the original CLIP, while maintaining richer semantic representations. The computational overhead is thus reduced to a one-time preprocessing step, enabling efficient training at scale.

\begin{table*}[t]
\small
\centering
\begin{NiceTabular}{l|l|cc|cc|cc|cc}
& & \multicolumn{2}{c}{S4V-val} & \multicolumn{2}{c}{Urban-1k} & \multicolumn{2}{c}{DCI} & \multicolumn{2}{c}{DOCCI-test} \\
Method & Dataset & I2T & T2I & I2T & T2I & I2T & T2I & I2T & T2I \\
\midrule[1pt]
CLIP~\cite{radford2021clip} & CC3M & 21.4 & 20.2 & 10.7 & 9.6 & 8.3 & 7.5 & 8.6 & 7.0 \\
Long-CLIP~\cite{zhang2024long-clip} & CC3M+S4V & 51.3 & 46.1 & 15.5 & 18.5 & 13.8 & 14.2 & 14.7 & 12.6 \\
FLAME & CC3M & \cellcolor{gray!10}\textbf{85.6} & \cellcolor{gray!10}\textbf{80.0} & \cellcolor{gray!10}\textbf{65.3} & \cellcolor{gray!10}\textbf{66.6} & \cellcolor{gray!10}\textbf{50.8} & \cellcolor{gray!10}\textbf{49.3} & \cellcolor{gray!10}\textbf{54.5} & \cellcolor{gray!10}\textbf{51.9} \\
\midrule
CLIP~\cite{radford2021clip} & YFCC15M & 57.1 & 45.9 & 30.4 & 23.6 & 22.0 & 19.1 & 26.5 & 23.3 \\
Long-CLIP~\cite{zhang2024long-clip} & YFCC15M+S4V & 77.2 & 77.6 & 40.5 & 46.1 & 29.4 & 29.7 & 35.0 & 33.5 \\
FLAME & YFCC15M & \cellcolor{gray!10}\textbf{94.1} & \cellcolor{gray!10}\textbf{93.2} & \cellcolor{gray!10}\textbf{84.0} & \cellcolor{gray!10}\textbf{87.9} & \cellcolor{gray!10}\textbf{66.1} & \cellcolor{gray!10}\textbf{68.1} & \cellcolor{gray!10}\textbf{75.8} & \cellcolor{gray!10}\textbf{76.2} \\
\midrule
CLIP~\cite{radford2021clip} & WIT-400M & 78.2 & 79.6 & 67.5 & 53.3 & 45.4 & 43.0 & 60.7 & 57.0 \\
\end{NiceTabular}
\vspace{-8pt}
\caption{\textbf{Long-context retrieval recall@1 results.} S4V denotes ShareGPT4V. FLAME trained on YFCC15M exhibits a remarkable 34.6\% improvement over WIT-400M-trained CLIP in text-to-image retrieval performance on Urban-1k.}
\label{tab:long_context_r}
\vspace{-1em}
\end{table*}
\section{Experiments}

\subsection{Datasets and Implementation Details}
\begin{figure}[t]
    \centering
    \includegraphics[width=0.99\linewidth]{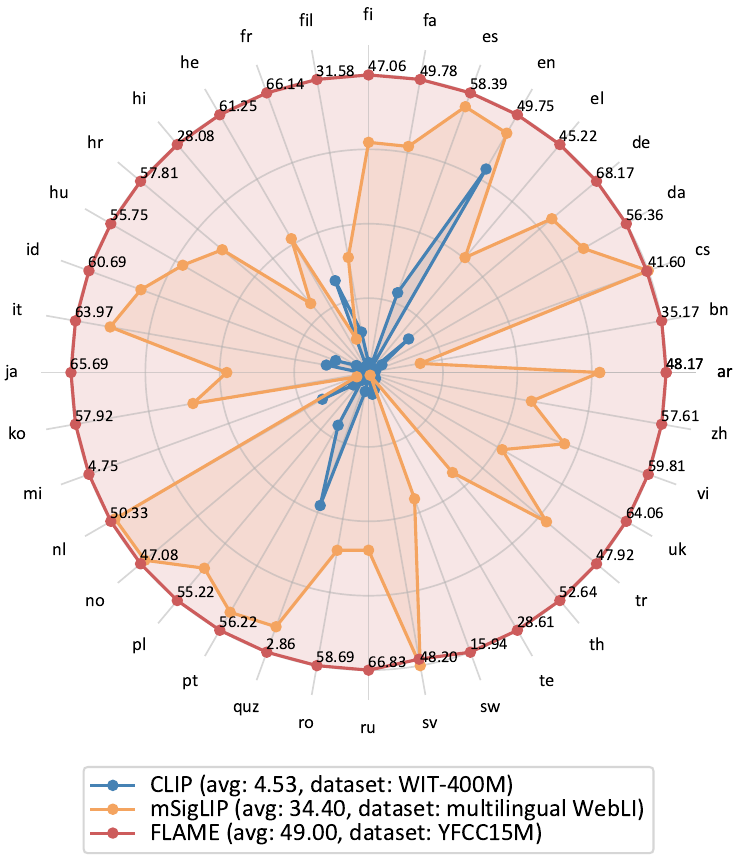}
\vspace{-8pt}
    \caption{\textbf{Multilingual zero-shot retrieval recall@1 results on Crossmodal-3600 (text-to-image retrieval).} Despite being trained solely on English datasets, FLAME achieves outstanding average performance across all 36 languages, surpassing mSigLIP, which is trained on the multilingual WebLI dataset with 100 languages. The image-to-text results are provided in the supplementary materials.}
    \label{fig:multilingual}
\vspace{-1.5em}
\end{figure}
\begin{table*}[t]
\small
\centering
\resizebox{0.9\linewidth}{!}{
\begin{NiceTabular}{l|l|cccccc|cccccc}
& & \multicolumn{6}{c}{Text Retrieval} & \multicolumn{6}{c}{Image Retrieval} \\
Dataset & Method & \multicolumn{3}{c}{MSCOCO} & \multicolumn{3}{c}{Flickr30k} & \multicolumn{3}{c}{MSCOCO} & \multicolumn{3}{c}{Flickr30k} \\
& & R@1 & R@5 & R@10 & R@1 & R@5 & R@10 & R@1 & R@5 & R@10 & R@1 & R@5 & R@10 \\
\midrule[1pt]
\multirow{4}{*}{CC3M} & CLIP~\cite{radford2021clip} & 8.7 & 23.9 & 33.7 & 7.1 & 19.7 & 28.6 & 17.4 & 37.9 & 50.1 & 13.9 & 30.8 & 40.5 \\
& MLLM-A~\cite{liu2023mllm-a} & 35.9 & 62.4 & 73.9 & 63.5 & 86.6 & 91.7 & 26.5 & 51.1 & 62.7 & 49.3 & 74.8 & 83.1 \\
& DreamLIP~\cite{zheng2024dreamlip} & \underline{39.9} & \underline{67.2} & \underline{78.1} & \underline{66.8} & \textbf{89.6} & \textbf{94.4} & \textbf{29.8} & \textbf{55.2} & \textbf{66.3} & \underline{50.7} & \underline{76.7} & \underline{83.6} \\
& FLAME & \cellcolor{gray!10}\textbf{43.3} & \cellcolor{gray!10}\textbf{69.1} & \cellcolor{gray!10}\textbf{78.9} & \cellcolor{gray!10}\textbf{67.3} & \cellcolor{gray!10}\underline{87.6} & \cellcolor{gray!10}\underline{93.1} & \cellcolor{gray!10}\underline{28.6} & \cellcolor{gray!10}\underline{54.5} & \cellcolor{gray!10}\underline{65.7} & \cellcolor{gray!10}\textbf{53.6} & \cellcolor{gray!10}\textbf{79.9} & \cellcolor{gray!10}\textbf{87.1} \\
\midrule
\multirow{4}{*}{YFCC15M} & CLIP~\cite{radford2021clip} & 30.7 & 56.2 & 67.4 & 54.9 & 80.0 & 88.4 & 19.1 & 40.9 & 52.5 & 37.2 & 64.3 & 74.3 \\
& SoftCLIP~\cite{gao2024softclip} & 30.9 & 56.2 & 68.3 & 56.2 & 82.1 & 88.6 & 19.2 & 41.2 & 52.6 & 37.2 & 64.3 & 74.5 \\
& DreamLIP~\cite{zheng2024dreamlip} & \underline{55.8} & \underline{80.7} & \underline{88.7} & \underline{84.9} & \textbf{97.3} & \textbf{99.1} & \underline{42.3} & \underline{68.9} & \underline{78.0} & \underline{65.3} & \underline{86.7} & \underline{91.8} \\
& FLAME & \cellcolor{gray!10}\textbf{60.5} & \cellcolor{gray!10}\textbf{82.9} & \cellcolor{gray!10}\textbf{89.3} & \cellcolor{gray!10}\textbf{86.4} & \cellcolor{gray!10}\textbf{97.3} & \cellcolor{gray!10}\underline{98.6} & \cellcolor{gray!10}\textbf{43.9} & \cellcolor{gray!10}\textbf{70.4} & \cellcolor{gray!10}\textbf{79.7} & \cellcolor{gray!10}\textbf{73.3} & \cellcolor{gray!10}\textbf{91.7} & \cellcolor{gray!10}\textbf{95.5} \\
\midrule
WIT-400M & CLIP~\cite{radford2021clip} & 52.4 & 76.7 & 84.7 & 81.9 & 96.2 & 98.8 & 33.1 & 58.4 & 69.0 & 62.1 & 85.6 & 91.8 \\
\end{NiceTabular}
}
\vspace{-8pt}
\caption{\textbf{Short-context retrieval.} The best results are in \textbf{bold} and the second best are \underline{underlined}. Our FLAME outperforms CLIP while using a significantly smaller dataset for training (YFCC15M v.s. WIT-400M).}
\label{tab:zs_r}
\vspace{-1.5em}
\end{table*}
\paragraph{Dataset Details.}  
We use two primary datasets for language-image pre-training: CC3M~\cite{sharma2018cc} and YFCC15M~\cite{li2021declip}. Our ablation studies are primarily conducted using the CC3M dataset. Given the differences in the availability of web-crawled data, we re-implement the data synthesis pipeline from DreamLIP~\cite{zheng2024dreamlip}, using MiniCPM-Llama3-V 2.5
to generate both long and short captions. The prompts are aligned with DreamLIP's approach, i.e., ``Describe the image in detail/short." The raw captions are also included during training. All these datasets will be released to the community to benefit other research. 

\paragraph{Implementation Details.}  
Our method is implemented using OpenCLIP\footnote{\url{https://github.com/mlfoundations/open_clip}}~\cite{openclip} and trained on a single node equipped with 8 NVIDIA A800 GPUs. The evaluation follows the CLIP benchmark protocol\footnote{\url{https://github.com/LAION-AI/CLIP_benchmark}} or the respective official benchmark protocols. We use ViT-B/16~\cite{dosovitskiy2020vit, radford2021clip, bai2024artrackv2} as the default visual encoder and Mistral-Nemo
~\cite{jiang2023mistral} as the default large language model (LLM). For tasks with long text input, we simply calculate the average cosine similarity between an image embedding and multiple text embeddings, avoiding complex methods. For tasks with short text input, we apply a single prompt for inference, which is the first prompt at the scene level provided in the supplementary material. For a fair comparison with previous methods, we train all models for 32 epochs. Input images are resized to 224 $\times$ 224. The learnable temperature parameter $\tau$ is initialized to 0.07. More detailed hyper-parameters are provided in the supplementary material. 

\subsection{Zero-shot Image-Text Retrieval}

\paragraph{Long-Context Retrieval.}  
FLAME excels in long-context image-text retrieval tasks, as demonstrated by evaluations on ShareGPT4V-val~\cite{chen2023sharegpt4v} and Urban-1k~\cite{zhang2024long-clip}. Despite being trained on YFCC15M, FLAME achieves an 87.9\% recall@1 in Urban-1k text-to-image retrieval, representing a 34.6\% improvement over CLIP trained on the much larger WIT-400M dataset. Additionally, FLAME demonstrates an average improvement of 14.8\% on ShareGPT4V-val. We also re-implement the recent method based on positional encoding interpolation~\cite{zhang2024long-clip}, i.e., pre-training a CLIP model on CC3M and YFCC15M, then using their official code
for fine-tuning on ShareGPT4V. The gap in results further demonstrates the advantage of FLAME’s paradigm shift over the interpolation-based architectural change on the standard CLIP text encoder.

\paragraph{Multilingual Retrieval.}  
FLAME leverages the intrinsic capabilities of pre-trained LLMs, enabling effective multilingual generalization. Although the model is trained on English datasets, it can be directly applied to downstream tasks in multiple languages without additional training. We evaluate this multilingual retrieval capability by conducting zero-shot retrieval across 36 languages from the Crossmodal-3600 dataset~\cite{thapliyal2022crossmodal-3600}. As shown in Figure~\ref{fig:multilingual} and Figure C1 in the supplementary materials, FLAME outperforms WIT-400M-trained CLIP~\cite{radford2021clip} across all 36 languages. Specifically, FLAME achieves an average image-to-text recall@1 of 51.7\% and text-to-image recall@1 of 49.0\%, surpassing CLIP by margins of 44.4\% and 44.5\%, respectively. Notably, FLAME outperforms mSigLIP~\cite{zhai2023siglip}, which is trained on the multilingual WebLI~\cite{chen2023pali} dataset with 100 languages, by 4.3\% in image-to-text recall@1 and 14.6\% in text-to-image recall@1.

\begin{table*}[ht]
\small
\centering
\resizebox{0.8\linewidth}{!}{
\begin{NiceTabular}{l|l|cccccccccc|c|c}
\rotatebox[origin=c]{0}{Dataset} & \rotatebox[origin=c]{0}{Method} & \rotatebox[origin=l]{90}{Food-101} & \rotatebox[origin=l]{90}{CIFAR-10} & \rotatebox[origin=l]{90}{CIFAR-100} &	\rotatebox[origin=l]{90}{SUN397} &	\rotatebox[origin=l]{90}{Cars} & \rotatebox[origin=l]{90}{Aircraft} & \rotatebox[origin=l]{90}{DTD} & \rotatebox[origin=l]{90}{Pets} & \rotatebox[origin=l]{90}{Caltech-101} & \rotatebox[origin=l]{90}{Flowers} & \rotatebox[origin=l]{90}{Average} & \rotatebox[origin=l]{90}{ImageNet} \\
\midrule[1pt]
\multirow{5}{*}{CC3M} & CLIP~\cite{radford2021clip} & 10.6 & 53.9 & 20.4 & 31.2 & 1.2 & 1.1 & 10.4 & 11.7 & 43.2 & 12.9 & 19.7 & 16.0 \\
& LaCLIP~\cite{fan2024laclip} & 14.2 & 57.1 & 27.5 & 35.1 & 1.6 & \underline{1.6} & 16.6 & 15.6 & 52.7 & 14.7 & 23.7 & 21.5 \\
& MLLM-A~\cite{liu2023mllm-a} & 18.7 & 58.4 & 32.4 & 43.8 & \underline{3.9} & 1.5 & \underline{20.2} & \underline{32.1} & \underline{63.5} & \underline{17.5} & 29.2 & 25.0 \\
& DreamLIP~\cite{zheng2024dreamlip} & \underline{19.4} & \textbf{74.3} & \textbf{44.2} & \underline{45.9} & 2.8 & 1.0 & 17.0 & 27.1 & 63.1 & 14.7 & \underline{31.0} & \underline{31.1} \\
& FLAME & \cellcolor{gray!10}\textbf{32.1} & \cellcolor{gray!10}\underline{73.6} & \cellcolor{gray!10}\underline{42.0} & \cellcolor{gray!10}\textbf{56.6} & \cellcolor{gray!10}\textbf{6.7} & \cellcolor{gray!10}\textbf{6.9} & \cellcolor{gray!10}\textbf{43.8} & \cellcolor{gray!10}\textbf{41.2} & \cellcolor{gray!10}\textbf{74.1} & \cellcolor{gray!10}\textbf{26.3} & \cellcolor{gray!10}\textbf{40.3} & \cellcolor{gray!10}\textbf{36.0} \\
\midrule
\multirow{3}{*}{YFCC15M} & CLIP~\cite{radford2021clip} & 35.0 & 67.1 & 34.8 & 42.0 & 5.1 & 6.3 & 13.9 & 20.4 & 54.5 & 44.3 & 32.3 & 34.1 \\
& DreamLIP~\cite{zheng2024dreamlip} & \underline{44.2} & \textbf{89.0} & \textbf{62.0} & \underline{57.1} & \underline{9.2} & \underline{6.4} & \underline{30.5} & \underline{32.6} & \textbf{79.8} & \underline{40.2} & \underline{45.1} & \underline{48.2} \\
& FLAME & \cellcolor{gray!10}\textbf{61.8} & \cellcolor{gray!10}\underline{86.1} & \cellcolor{gray!10}\underline{56.7} & \cellcolor{gray!10}\textbf{66.8} & \cellcolor{gray!10}\textbf{10.7} & \cellcolor{gray!10}\textbf{10.3} & \cellcolor{gray!10}\textbf{54.9} & \cellcolor{gray!10}\textbf{40.7} & \cellcolor{gray!10}\underline{78.9} & \cellcolor{gray!10}\textbf{51.7} & \cellcolor{gray!10}\textbf{51.9} & \cellcolor{gray!10}\textbf{51.5} \\
\end{NiceTabular}
}
\vspace{-8pt}
\caption{\textbf{Zero-shot classification.} FLAME significantly improves top-1 accuracy on ImageNet and average top-1 accuracy on 10 downstream datasets over the previous methods.}
\label{tab:zs_cls}
\vspace{-2em}
\end{table*}
\begin{table}[h]
\small
\centering
\resizebox{\linewidth}{!}{
\begin{NiceTabular}{l|l|cccccc}
& & \multicolumn{6}{c}{Evaluation} \\
Method & Training & ar & en & it & jp & zh & avg \\
\midrule[1pt]
CLIP~\cite{radford2021clip} & en & 0.4 & \textbf{68.3} & 21.8 & 4.2 & 1.4 & 19.2 \\
CN-CLIP~\cite{yang2022cn-clip} & en+zh & 0.1 & 27.7 & 7.5 & 14.6 & \textbf{47.0} & 19.4 \\
NLLB-CLIP~\cite{visheratin2023nllb-clip} & en+multi & 18.4 & 24.3 & 18.0 & 16.5 & 16.6 & 18.8 \\
FLAME & en & \cellcolor{gray!10}\textbf{32.7} & \cellcolor{gray!10}51.5 & \cellcolor{gray!10}\textbf{42.1} & \cellcolor{gray!10}\textbf{36.9} & \cellcolor{gray!10}39.7 & \cellcolor{gray!10}\textbf{40.6} \\

\end{NiceTabular}
}
\vspace{-8pt}
\caption{\textbf{Multilingual ImageNet classification.} CLIP is trained on WIT-400M. CN-CLIP is pre-trained on WIT-400M and fine-tuned on 200M Chinese data. NLLB-CLIP is pre-trained on WIT-400M and fine-tuned on LAION-COCO-NLLB with 200 languages. FLAME is trained on YFCC15M. Unlike other methods, FLAME generalizes from monolingual training to multilingual evaluation.}
\label{tab:multilingual_cls}
\vspace{-1em}
\end{table}

\paragraph{Short-Context Retrieval.}  
FLAME also significantly outperforms CLIP in short-context retrieval tasks. When trained on YFCC15M, it shows an 8.7\%/4.4\% average improvement on MSCOCO/Flickr30k compared to CLIP trained on WIT-400M, as summarized in Table~\ref{tab:zs_r}. These results indicate that while FLAME is designed for fine-grained semantic learning of long texts, it also achieves competent representations for context-limited texts, highlighting its potential in more general downstream scenarios. 

\subsection{Image Classification}

\paragraph{Zero-Shot Classification.}  
We evaluate FLAME on zero-shot image classification tasks across diverse datasets. As shown in Table~\ref{tab:zs_cls}, FLAME achieves a 4.9\% and 3.3\% improvement in ImageNet top-1 accuracy when trained on CC3M and YFCC15M, respectively. On 10 common downstream datasets, FLAME achieves an average accuracy of 40.3\%/51.9\%, improving by 9.3\%/6.8\% over the prior state-of-the-art~\cite{zheng2024dreamlip}. These results highlight that FLAME obtains powerful classification capabilities by aligning fine-grained entity-level representations.

\paragraph{Multilingual Classification.}  
To evaluate multilingual capabilities in image classification of FLAME, we conduct experiments on the multilingual ImageNet1k benchmark, covering Arabic (ar), English (en), Italian (it), Japanese (ja), and Chinese (zh). Results in Table~\ref{tab:multilingual_cls} show that while CLIP trained on WIT-400M performs well in English, it lacks multilingual performance. We find that models like CN-CLIP~\cite{yang2022cn-clip} trained on specific languages compromise their performance in others. In contrast, FLAME generalizes effectively across all five languages, maintaining strong performance even when trained only on an English dataset. Additionally, our results comprehensively outperform models trained on multilingual datasets~\cite{visheratin2023nllb-clip}, with an average accuracy improvement of 21.8\%.

\paragraph{Linear-Probe Classification.}  
We further evaluate FLAME's feature extraction capabilities through linear-probe experiments, as shown in Table~\ref{tab:lp_cls}. On downstream datasets, FLAME improves by 3.4\% and 3.0\% on average when trained on CC3M and YFCC15M, respectively. These results demonstrate that aligning visual encoders with LLMs enhances generalization across downstream tasks. 

\begin{table}[t]
\small
\centering
\resizebox{\linewidth}{!}{
\begin{NiceTabular}{l|l|ccccccc|c}
\rotatebox[origin=c]{0}{Dataset} & \rotatebox[origin=c]{0}{Method} & \rotatebox[origin=l]{90}{Food-101} & \rotatebox[origin=l]{90}{CIFAR-10} & \rotatebox[origin=l]{90}{CIFAR-100} &	\rotatebox[origin=l]{90}{Cars} & \rotatebox[origin=l]{90}{Aircraft} & \rotatebox[origin=l]{90}{DTD} & \rotatebox[origin=l]{90}{Caltech-101} & \rotatebox[origin=l]{90}{Average} \\
\midrule[1pt]
\multirow{5}{*}{CC3M} & CLIP~\cite{radford2021clip} & 62.6 & 86.8 & 68.1 & 32.8 & 40.9 & 63.4 & 82.0 & 62.4 \\
& LaCLIP~\cite{fan2024laclip} & 63.8 & 87.7 & 69.5 & 32.4 & \textbf{42.7} & 64.0 & 83.3 & 63.3 \\
& MLLM-A~\cite{liu2023mllm-a} & 64.0 & 87.7 & 68.5 & \underline{34.5} & 32.1 & 60.4 & 85.5 & 61.8 \\
& DreamLIP~\cite{zheng2024dreamlip} & \underline{71.2} & \textbf{92.2} & \textbf{74.0} & 31.5 & 26.7 & \textbf{70.4} & \underline{88.5} & \underline{64.9} \\
& FLAME & \cellcolor{gray!10}\textbf{72.0} & \cellcolor{gray!10}\underline{90.3} & \cellcolor{gray!10}\underline{72.9} & \cellcolor{gray!10}\textbf{45.2} & \cellcolor{gray!10}\underline{38.6} & \cellcolor{gray!10}\underline{69.7} & \cellcolor{gray!10}\textbf{89.5} & \cellcolor{gray!10}\textbf{68.3} \\
\midrule
\multirow{4}{*}{YFCC15M} & CLIP~\cite{radford2021clip} & 77.2 & 88.5 & 66.4 & 29.0 & 25.5 & 65.2 & 82.4 & 62.0 \\
& HiCLIP~\cite{geng2022hiclip} & 81.0 & 89.1 & 70.4 & 36.4 & 32.3 & 68.7 & 86.4 & 66.3 \\
& DreamLIP~\cite{zheng2024dreamlip} & \underline{83.6} & \textbf{96.5} & \textbf{82.3} & \underline{41.8} & \underline{34.6} & \underline{74.3} & \underline{91.2} & \underline{72.0} \\
& FLAME & \cellcolor{gray!10}\textbf{85.9} & \cellcolor{gray!10}\underline{95.0} & \cellcolor{gray!10}\underline{81.0} & \cellcolor{gray!10}\textbf{54.3} & \cellcolor{gray!10}\textbf{39.3} & \cellcolor{gray!10}\textbf{76.8} & \cellcolor{gray!10}\textbf{92.5} & \cellcolor{gray!10}\textbf{75.0} \\
\end{NiceTabular}
}
\vspace{-8pt}
\caption{\textbf{Linear-probe classification.} FLAME achieves competent enhancements of average accuracy across downstream datasets.}
\label{tab:lp_cls}
\vspace{-1.5em}
\end{table}

\begin{figure}[t]
\centering
\includegraphics[width=0.9\linewidth]{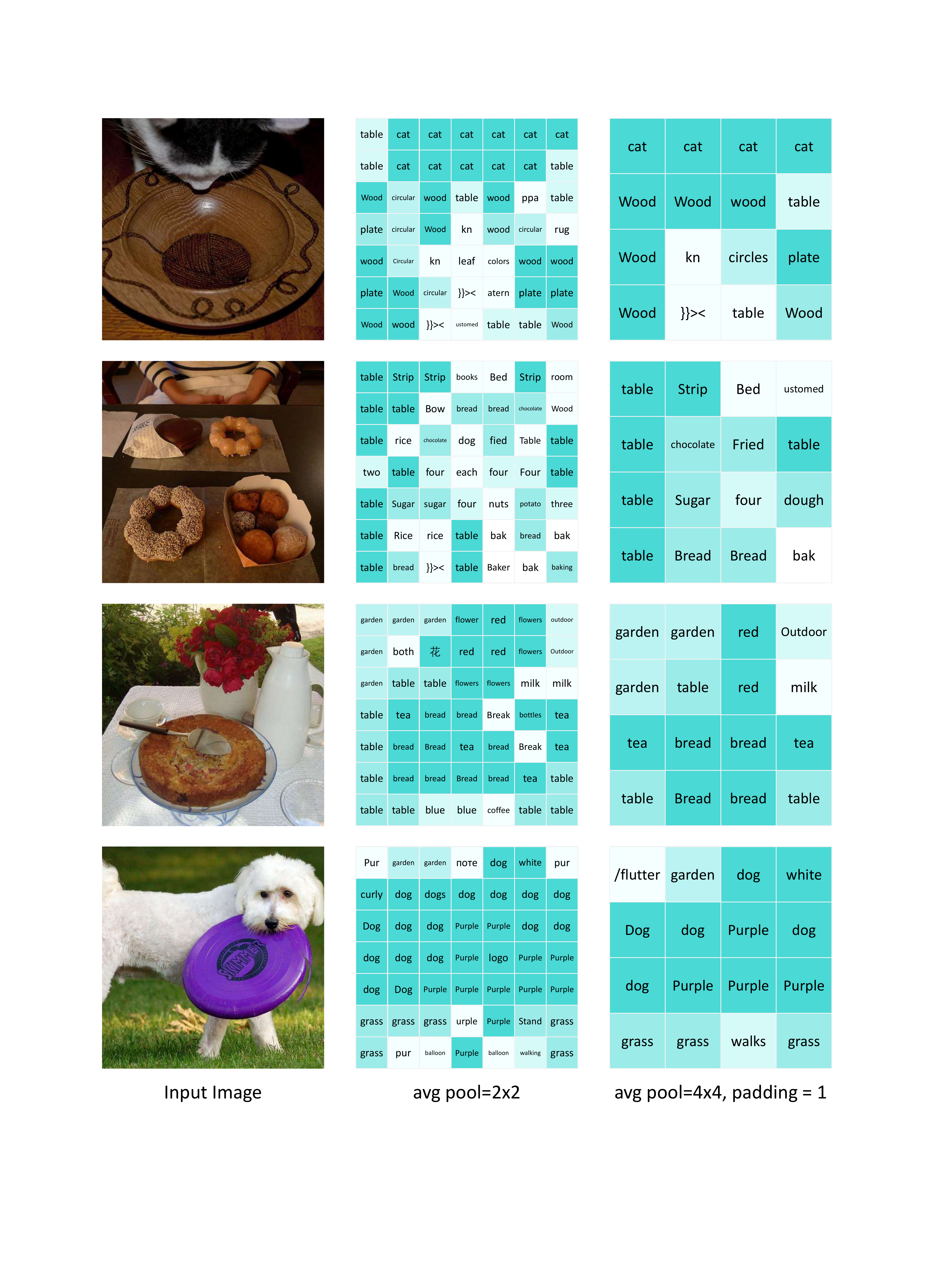}
    \vspace{-8pt}
    \caption{\textbf{Semantic interpretability.} Based on vocabulary mapping, FLAME achieves patch-to-word translation with competent interpretability of language-image alignment. We apply average pooling to reduce the number of words for a clearer presentation.}
    \label{fig:semantic_interpretability}
    \vspace{-1.5em}
\end{figure}

\begin{table*}[t]
\small
\centering
\resizebox{0.8\linewidth}{!}{
\begin{NiceTabular}{c|cc|cc|cc|cc|cc}
& \multicolumn{2}{c}{ShareGPT4V-val} & \multicolumn{2}{c}{Urban-1k} & \multicolumn{2}{c}{MSCOCO} & \multicolumn{2}{c}{Flickr30k} & \multicolumn{2}{c}{Classfication} \\
\# Prompts & I2T@1 & T2I@1 & I2T@1 & T2I@1 & I2T@1 & T2I@1 & I2T@1 & T2I@1 & ImageNet & DS Avg. \\
\midrule[1pt]
1 & 68.2 & 65.0 & 45.7 & 41.3 & 35.7 & 23.7 & 61.0 & 47.4 & 33.8 & 38.5 \\
3 & 72.7 & 68.5 & 53.9 & 46.0 & 37.4 & 25.1 & 62.5 & 49.9 & 34.1 & 39.0 \\
5 & 79.5 & 74.1 & 59.8 & 57.0 & 40.0 & 26.9 & 65.1 & 52.2 & 35.2 & 39.9 \\
7 & \cellcolor{gray!10}\textbf{85.6} & \cellcolor{gray!10}80.0 & \cellcolor{gray!10}65.3 & \cellcolor{gray!10}66.6 & \cellcolor{gray!10}\textbf{43.3} & \cellcolor{gray!10}\textbf{28.6} & \cellcolor{gray!10}67.3 & \cellcolor{gray!10}\textbf{54.6} & \cellcolor{gray!10}\textbf{36.0} & \cellcolor{gray!10}\textbf{40.3} \\
9 & 83.8 & \textbf{80.8} & \textbf{67.1} & \textbf{67.1} & 41.6 & 28.5 & \textbf{68.5} & 54.2 & 35.6 & 40.0 \\
\end{NiceTabular}
}
\vspace{-8pt}
\caption{\textbf{Ablations on the number of prompts.} Leveraging multifaceted prompts enhances performance, particularly in terms of long-context benchmarks. The right amount of prompts can achieve the best-balanced performance.}
\label{tab:ablation_prompt}
\vspace{-2em}
\end{table*}

\begin{table}[t]
\small
\centering
\begin{NiceTabular}{cl|cc}
& Backbone & ImageNet & DS Avg. \\
\midrule[1pt]
\multirow{3}{*}{\rotatebox[origin=c]{90}{Text}} & Mistral-7B & 34.4 & 39.6 \\ 
& Mistral-Nemo & \cellcolor{gray!10}\textbf{36.0} & \cellcolor{gray!10}\textbf{40.3} \\
& Llama-3.1-8B & 34.1 & 39.8 \\
\midrule
\multirow{3}{*}{\rotatebox[origin=c]{90}{Visual}} & ViT-S/16 & 32.4 (+19.0)  & 37.1 (+20.7) \\
& ViT-B/16 & \cellcolor{gray!10}36.0 (+20.0) & \cellcolor{gray!10}40.3 (+20.6) \\
& ViT-L/14 & \textbf{38.7 (+22.5)} & \textbf{42.9 (+22.2)} \\
\end{NiceTabular}
\vspace{-8pt}
\caption{\textbf{Ablations on backbone architectures.} The values in parentheses indicate improvements over the CLIP baseline using the same visual backbone. Whether employing larger or compact architectures, our framework demonstrates sufficient robustness.}
\label{tab:ablation_backbone}
\vspace{-1.5em}
\end{table}

\subsection{Semantic Interpretability} 
\label{subsec:semantic_interpretability}
Since we align the visual encoder with the last hidden layer of the LLM, we can input each visual output into the projection layer of the LLM to get the next-token logits. We then look up the token with the highest probability in the LLM's vocabulary, allowing each image patch to be semantically interpreted as a word. The overall process can be expressed as: $C_i = M(h(f_I(x_i)))$, where $h$ denotes the projection head of the LLM, $M$ represents the mapping process that identifies the word with the highest probability in the LLM's vocabulary. As shown in Figure~\ref{fig:semantic_interpretability}, the mapped words reflect meaningful semantic properties such as categories (e.g., cat, plate), attributes (e.g., circular), and numerical values (e.g., four). This semantic mapping enhances interpretability by translating image features into understandable text.

\subsection{Ablation Studies}

\paragraph{Number of Prompts.} The quantity of prompts used plays a crucial role in the granularity of information extracted from long texts. To explore this relationship, we conducted an extensive ablation study on the number of prompts. The evaluation covers long/short-context retrieval and image classification tasks, with the results summarized in Table~\ref{tab:ablation_prompt}. DS refers to the downstream datasets from Table~\ref{tab:zs_cls}. We began with a diverse set of 9 prompts and performed random sampling to employ varying numbers of prompts for experimentation.

Our results show that using too few prompts leads to suboptimal performance, particularly on the long-context benchmarks. This phenomenon is due to the incomplete extraction and learning of features from long captions. Notably, the performance drop is less pronounced on classification tasks involving simpler short text inputs. On the other hand, when more than 7 prompts are used, the accuracy improvement plateaus.
Therefore, we use 7 as the default number of prompts to achieve a more balanced performance.

\paragraph{Different Backbone Architectures.} To assess the robustness and scalability of our framework, we conducted ablation studies using different backbone architectures. These included various text encoders (Mistral-7B
, Mistral-Nemo, and Llama-3.1-8B
) and visual encoders (ViT-S/16, ViT-B/16, and ViT-L/14). Whether reducing or increasing the architecture size, our framework continues to perform robustly. For instance, when using Mistral-7B, our model achieves an ImageNet top-1 accuracy of 34.4\% and an average accuracy of 39.6\% on downstream datasets, both of which significantly surpass the prior state-of-the-art results (31.1\% and 31.0\%, respectively)~\cite{zheng2024dreamlip}. Additionally, we observe that performance improvements scale with the size of the visual encoder. For instance, FLAME with the ViT-L/14 encoder increases ImageNet top-1 accuracy by 22.5\% over the CLIP baseline, while using ViT-S/16 yields a 19.0\% improvement, underscoring the scalability of our approach.

\begin{figure}[t]
\centering
\includegraphics[width=0.9\linewidth]{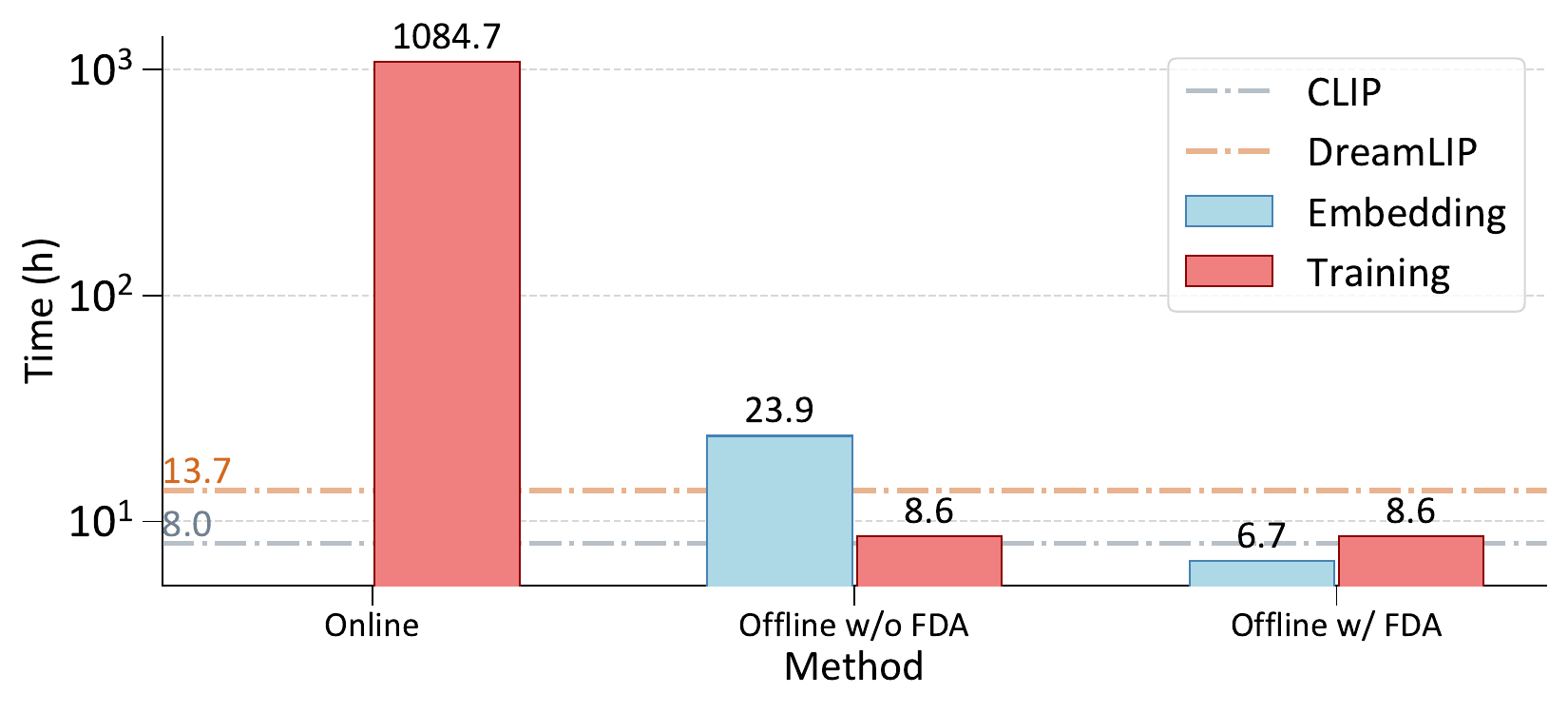}
    \vspace{-12pt}
    \caption{\textbf{Embedding and training overhead.} FDA denotes facet-decoupled attention. While online training is hindered by inefficiencies, our approach of freezing LLMs allows for offline embedding to lower the overhead. Our FDA further boosts the efficiency of offline embedding, resulting in a 3.6x speedup.}
    \label{fig:ablation_overhead}
    \vspace{-1.5em}
\end{figure}

\paragraph{Embedding and Training Overhead.} We evaluate the embedding and training overhead on CC3M, with results illustrated in Figure~\ref{fig:ablation_overhead}. For the recent method~\cite{zheng2024dreamlip}, we utilize their official code
to measure training time. As shown, online training suffers from inefficiencies due to the frequent forward passes of LLMs. Our option of freezing LLMs, however, enables offline embedding preprocessing, significantly reducing unnecessary overhead. Furthermore, our proposed facet-decoupled attention greatly enhances the efficiency of offline embedding, achieving a 3.6x speedup. Our training time remains competitive with CLIP and outperforms that of the recent method~\cite{zheng2024dreamlip}.

\section{Conclusion and Limitation}
In this paper, we introduced FLAME, a novel language-image pre-training framework that harnesses frozen large language models as text encoders. Through multifaceted prompt distillation and facet-decoupled attention mechanisms, FLAME demonstrates superior performance across long-context understanding, multilingual generalization, and data-efficient visual-semantic alignment. While the framework introduces additional computational overhead during the initial text embedding extraction phase, our attention optimization achieves more than 3.6x speedup compared to naive implementation, making the preprocessing cost manageable and worthwhile given the significant performance gains. With the rapid advancement of more efficient LLMs and the framework's demonstrated scalability across model sizes, FLAME represents a promising direction for future large-scale language-image pre-training.

\section*{Acknowledgements}
This work was supported in part by the National Natural Science Foundation of China (62206271), the Fundamental Research Funds for the Central Universities No.xxj032023020, the Shenzhen Key Technical Projects under Grant (JSGG20220831105801004, JCYJ20220818101406014, CJGJZD2022051714160501), the Guangdong University Featured Innovation Program Project (2024KTSCX026), and the CAAI-MindSpore Open Fund, developed on OpenI Community.

{
    \small
    \bibliographystyle{ieeenat_fullname}
    \bibliography{main}

\begin{thebibliography}{65}
\providecommand{\natexlab}[1]{#1}
\providecommand{\url}[1]{\texttt{#1}}
\expandafter\ifx\csname urlstyle\endcsname\relax
  \providecommand{\doi}[1]{doi: #1}\else
  \providecommand{\doi}{doi: \begingroup \urlstyle{rm}\Url}\fi

\bibitem[Abdin et~al.(2024)Abdin, Jacobs, Awan, Aneja, Awadallah, Awadalla, Bach, Bahree, Bakhtiari, Behl, et~al.]{abdin2024phi3}
Marah Abdin, Sam~Ade Jacobs, Ammar~Ahmad Awan, Jyoti Aneja, Ahmed Awadallah, Hany Awadalla, Nguyen Bach, Amit Bahree, Arash Bakhtiari, Harkirat Behl, et~al.
\newblock Phi-3 technical report: A highly capable language model locally on your phone.
\newblock \emph{arXiv preprint arXiv:2404.14219}, 2024.

\bibitem[Achiam et~al.(2023)Achiam, Adler, Agarwal, Ahmad, Akkaya, Aleman, Almeida, Altenschmidt, Altman, Anadkat, et~al.]{achiam2023gpt4}
Josh Achiam, Steven Adler, Sandhini Agarwal, Lama Ahmad, Ilge Akkaya, Florencia~Leoni Aleman, Diogo Almeida, Janko Altenschmidt, Sam Altman, Shyamal Anadkat, et~al.
\newblock Gpt-4 technical report.
\newblock \emph{arXiv preprint arXiv:2303.08774}, 2023.

\bibitem[Bai et~al.(2023)Bai, Bai, Chu, Cui, Dang, Deng, Fan, Ge, Han, Huang, et~al.]{bai2023qwen}
Jinze Bai, Shuai Bai, Yunfei Chu, Zeyu Cui, Kai Dang, Xiaodong Deng, Yang Fan, Wenbin Ge, Yu Han, Fei Huang, et~al.
\newblock Qwen technical report.
\newblock \emph{arXiv preprint arXiv:2309.16609}, 2023.

\bibitem[Bai et~al.(2024{\natexlab{a}})Bai, Wu, Liu, Jia, Mao, Zhang, Zhao, Shen, Wei, Wang, et~al.]{bai2024atlas}
Yifan Bai, Dongming Wu, Yingfei Liu, Fan Jia, Weixin Mao, Ziheng Zhang, Yucheng Zhao, Jianbing Shen, Xing Wei, Tiancai Wang, et~al.
\newblock Is a 3d-tokenized llm the key to reliable autonomous driving?
\newblock \emph{arXiv preprint arXiv:2405.18361}, 2024{\natexlab{a}}.

\bibitem[Bai et~al.(2024{\natexlab{b}})Bai, Zhao, Gong, and Wei]{bai2024artrackv2}
Yifan Bai, Zeyang Zhao, Yihong Gong, and Xing Wei.
\newblock Artrackv2: Prompting autoregressive tracker where to look and how to describe.
\newblock In \emph{CVPR}, 2024{\natexlab{b}}.

\bibitem[BehnamGhader et~al.(2024)BehnamGhader, Adlakha, Mosbach, Bahdanau, Chapados, and Reddy]{behnamghader2024llm2vec}
Parishad BehnamGhader, Vaibhav Adlakha, Marius Mosbach, Dzmitry Bahdanau, Nicolas Chapados, and Siva Reddy.
\newblock Llm2vec: Large language models are secretly powerful text encoders.
\newblock \emph{arXiv preprint arXiv:2404.05961}, 2024.

\bibitem[Brown et~al.(2020)Brown, Mann, Ryder, Subbiah, Kaplan, Dhariwal, Neelakantan, Shyam, Sastry, Askell, et~al.]{brown2020gpt3}
Tom Brown, Benjamin Mann, Nick Ryder, Melanie Subbiah, Jared~D Kaplan, Prafulla Dhariwal, Arvind Neelakantan, Pranav Shyam, Girish Sastry, Amanda Askell, et~al.
\newblock Language models are few-shot learners.
\newblock In \emph{NeurIPS}, 2020.

\bibitem[Carlsson et~al.(2022)Carlsson, Eisen, Rekathati, and Sahlgren]{carlsson2022mclip}
Fredrik Carlsson, Philipp Eisen, Faton Rekathati, and Magnus Sahlgren.
\newblock Cross-lingual and multilingual clip.
\newblock In \emph{LREC}, 2022.

\bibitem[Chen et~al.(2023{\natexlab{a}})Chen, Li, Dong, Zhang, He, Wang, Zhao, and Lin]{chen2023sharegpt4v}
Lin Chen, Jinsong Li, Xiaoyi Dong, Pan Zhang, Conghui He, Jiaqi Wang, Feng Zhao, and Dahua Lin.
\newblock Sharegpt4v: Improving large multi-modal models with better captions.
\newblock \emph{arXiv preprint arXiv:2311.12793}, 2023{\natexlab{a}}.

\bibitem[Chen et~al.(2023{\natexlab{b}})Chen, Wang, Changpinyo, Piergiovanni, Padlewski, Salz, Goodman, Grycner, Mustafa, Beyer, et~al.]{chen2023pali}
Xi Chen, Xiao Wang, Soravit Changpinyo, AJ Piergiovanni, Piotr Padlewski, Daniel Salz, Sebastian Goodman, Adam Grycner, Basil Mustafa, Lucas Beyer, et~al.
\newblock Pali: A jointly-scaled multilingual language-image model.
\newblock In \emph{ICLR}, 2023{\natexlab{b}}.

\bibitem[Chen et~al.(2023{\natexlab{c}})Chen, Liu, Zhang, Yang, and Wu]{chen2023altclip}
Zhongzhi Chen, Guang Liu, Bo-Wen Zhang, Qinghong Yang, and Ledell Wu.
\newblock Altclip: Altering the language encoder in clip for extended language capabilities.
\newblock In \emph{ACL Findings}, 2023{\natexlab{c}}.

\bibitem[Dong et~al.(2023)Dong, Bao, Zheng, Zhang, Chen, Yang, Zeng, Zhang, Yuan, Chen, et~al.]{dong2023maskclip}
Xiaoyi Dong, Jianmin Bao, Yinglin Zheng, Ting Zhang, Dongdong Chen, Hao Yang, Ming Zeng, Weiming Zhang, Lu Yuan, Dong Chen, et~al.
\newblock Maskclip: Masked self-distillation advances contrastive language-image pretraining.
\newblock In \emph{CVPR}, 2023.

\bibitem[Dosovitskiy et~al.(2021)Dosovitskiy, Beyer, Kolesnikov, Weissenborn, Zhai, Unterthiner, Dehghani, Minderer, Heigold, Gelly, et~al.]{dosovitskiy2020vit}
Alexey Dosovitskiy, Lucas Beyer, Alexander Kolesnikov, Dirk Weissenborn, Xiaohua Zhai, Thomas Unterthiner, Mostafa Dehghani, Matthias Minderer, Georg Heigold, Sylvain Gelly, et~al.
\newblock An image is worth 16x16 words: Transformers for image recognition at scale.
\newblock In \emph{ICLR}, 2021.

\bibitem[Fan et~al.(2023)Fan, Krishnan, Isola, Katabi, and Tian]{fan2024laclip}
Lijie Fan, Dilip Krishnan, Phillip Isola, Dina Katabi, and Yonglong Tian.
\newblock Improving clip training with language rewrites.
\newblock In \emph{NeurIPS}, 2023.

\bibitem[Gao et~al.(2022)Gao, Liu, Xu, Zhang, Li, Ji, and Shen]{gao2022pyramidclip}
Yuting Gao, Jinfeng Liu, Zihan Xu, Jun Zhang, Ke Li, Rongrong Ji, and Chunhua Shen.
\newblock Pyramidclip: Hierarchical feature alignment for vision-language model pretraining.
\newblock In \emph{NeurIPS}, 2022.

\bibitem[Gao et~al.(2024)Gao, Liu, Xu, Wu, Zhang, Li, Yang, Liu, and Sun]{gao2024softclip}
Yuting Gao, Jinfeng Liu, Zihan Xu, Tong Wu, Enwei Zhang, Ke Li, Jie Yang, Wei Liu, and Xing Sun.
\newblock Softclip: Softer cross-modal alignment makes clip stronger.
\newblock In \emph{AAAI}, 2024.

\bibitem[Geng et~al.(2023)Geng, Yuan, Tian, Chen, and Zhang]{geng2022hiclip}
Shijie Geng, Jianbo Yuan, Yu Tian, Yuxiao Chen, and Yongfeng Zhang.
\newblock Hiclip: Contrastive language-image pretraining with hierarchy-aware attention.
\newblock In \emph{ICLR}, 2023.

\bibitem[Hsieh et~al.(2023)Hsieh, Zhang, Ma, Kembhavi, and Krishna]{hsieh2024sugarcrepe}
Cheng-Yu Hsieh, Jieyu Zhang, Zixian Ma, Aniruddha Kembhavi, and Ranjay Krishna.
\newblock Sugarcrepe: Fixing hackable benchmarks for vision-language compositionality.
\newblock In \emph{NeurIPS}, 2023.

\bibitem[Ilharco et~al.(2021)Ilharco, Wortsman, Wightman, Gordon, Carlini, Taori, Dave, Shankar, Namkoong, Miller, Hajishirzi, Farhadi, and Schmidt]{openclip}
Gabriel Ilharco, Mitchell Wortsman, Ross Wightman, Cade Gordon, Nicholas Carlini, Rohan Taori, Achal Dave, Vaishaal Shankar, Hongseok Namkoong, John Miller, Hannaneh Hajishirzi, Ali Farhadi, and Ludwig Schmidt.
\newblock Openclip, 2021.
\newblock If you use this software, please cite it as below.

\bibitem[Jia et~al.(2021)Jia, Yang, Xia, Chen, Parekh, Pham, Le, Sung, Li, and Duerig]{jia2021align}
Chao Jia, Yinfei Yang, Ye Xia, Yi-Ting Chen, Zarana Parekh, Hieu Pham, Quoc Le, Yun-Hsuan Sung, Zhen Li, and Tom Duerig.
\newblock Scaling up visual and vision-language representation learning with noisy text supervision.
\newblock In \emph{ICML}, 2021.

\bibitem[Jiang et~al.(2023{\natexlab{a}})Jiang, Sablayrolles, Mensch, Bamford, Chaplot, Casas, Bressand, Lengyel, Lample, Saulnier, et~al.]{jiang2023mistral}
Albert~Q Jiang, Alexandre Sablayrolles, Arthur Mensch, Chris Bamford, Devendra~Singh Chaplot, Diego de~las Casas, Florian Bressand, Gianna Lengyel, Guillaume Lample, Lucile Saulnier, et~al.
\newblock Mistral 7b.
\newblock \emph{arXiv preprint arXiv:2310.06825}, 2023{\natexlab{a}}.

\bibitem[Jiang et~al.(2023{\natexlab{b}})Jiang, Huang, Luan, Wang, and Zhuang]{jiang2023prompteol}
Ting Jiang, Shaohan Huang, Zhongzhi Luan, Deqing Wang, and Fuzhen Zhuang.
\newblock Scaling sentence embeddings with large language models.
\newblock \emph{arXiv preprint arXiv:2307.16645}, 2023{\natexlab{b}}.

\bibitem[Kim et~al.(2023)Kim, Jo, Kim, and Kim]{kim2023mcd}
Bumsoo Kim, Yeonsik Jo, Jinhyung Kim, and Seunghwan Kim.
\newblock Misalign, contrast then distill: Rethinking misalignments in language-image pre-training.
\newblock In \emph{ICCV}, 2023.

\bibitem[Koukounas et~al.(2024)Koukounas, Mastrapas, G{\"u}nther, Wang, Martens, Mohr, Sturua, Akram, Mart{\'\i}nez, Ognawala, et~al.]{koukounas2024jinaclip}
Andreas Koukounas, Georgios Mastrapas, Michael G{\"u}nther, Bo Wang, Scott Martens, Isabelle Mohr, Saba Sturua, Mohammad~Kalim Akram, Joan~Fontanals Mart{\'\i}nez, Saahil Ognawala, et~al.
\newblock Jina clip: Your clip model is also your text retriever.
\newblock \emph{arXiv preprint arXiv:2405.20204}, 2024.

\bibitem[Lai et~al.(2023)Lai, Zhang, Wu, Bai, Timofeev, Du, Gan, Shan, Chuah, Yang, et~al.]{lai2023veclip}
Zhengfeng Lai, Haotian Zhang, Wentao Wu, Haoping Bai, Aleksei Timofeev, Xianzhi Du, Zhe Gan, Jiulong Shan, Chen-Nee Chuah, Yinfei Yang, et~al.
\newblock From scarcity to efficiency: Improving clip training via visual-enriched captions.
\newblock \emph{arXiv preprint arXiv:2310.07699}, 2023.

\bibitem[Lee et~al.(2022)Lee, Kim, Shon, Kim, Kim, Lee, and Kim]{lee2022uniclip}
Janghyeon Lee, Jongsuk Kim, Hyounguk Shon, Bumsoo Kim, Seung~Hwan Kim, Honglak Lee, and Junmo Kim.
\newblock Uniclip: Unified framework for contrastive language-image pre-training.
\newblock In \emph{NeurIPS}, 2022.

\bibitem[Lee et~al.(2024)Lee, Dai, Ren, Chen, Cer, Cole, Hui, Boratko, Kapadia, Ding, et~al.]{lee2024gecko}
Jinhyuk Lee, Zhuyun Dai, Xiaoqi Ren, Blair Chen, Daniel Cer, Jeremy~R Cole, Kai Hui, Michael Boratko, Rajvi Kapadia, Wen Ding, et~al.
\newblock Gecko: Versatile text embeddings distilled from large language models.
\newblock \emph{arXiv preprint arXiv:2403.20327}, 2024.

\bibitem[Lei et~al.(2024)Lei, Wu, Zhou, Shen, Cao, Tao, and Yates]{lei2024metaeol}
Yibin Lei, Di Wu, Tianyi Zhou, Tao Shen, Yu Cao, Chongyang Tao, and Andrew Yates.
\newblock Meta-task prompting elicits embedding from large language models.
\newblock In \emph{ACL}, 2024.

\bibitem[Li et~al.(2024)Li, Zhu, Zhang, Jiang, Dang, Hou, Shen, Zhao, Shah, and Bennamoun]{li2024scene}
Hongsheng Li, Guangming Zhu, Liang Zhang, Youliang Jiang, Yixuan Dang, Haoran Hou, Peiyi Shen, Xia Zhao, Syed Afaq~Ali Shah, and Mohammed Bennamoun.
\newblock Scene graph generation: A comprehensive survey.
\newblock \emph{Neurocomputing}, 566:\penalty0 127052, 2024.

\bibitem[Li et~al.(2022{\natexlab{a}})Li, Zhang, Zhang, Yang, Li, Zhong, Wang, Yuan, Zhang, Hwang, et~al.]{li2022glip}
Liunian~Harold Li, Pengchuan Zhang, Haotian Zhang, Jianwei Yang, Chunyuan Li, Yiwu Zhong, Lijuan Wang, Lu Yuan, Lei Zhang, Jenq-Neng Hwang, et~al.
\newblock Grounded language-image pre-training.
\newblock In \emph{CVPR}, 2022{\natexlab{a}}.

\bibitem[Li et~al.(2023{\natexlab{a}})Li, Wang, and Xie]{li2023clipa}
Xianhang Li, Zeyu Wang, and Cihang Xie.
\newblock An inverse scaling law for clip training.
\newblock In \emph{NeurIPS}, 2023{\natexlab{a}}.

\bibitem[Li et~al.(2017)Li, Ouyang, Zhou, Wang, and Wang]{li2017scene}
Yikang Li, Wanli Ouyang, Bolei Zhou, Kun Wang, and Xiaogang Wang.
\newblock Scene graph generation from objects, phrases and region captions.
\newblock In \emph{ICCV}, 2017.

\bibitem[Li et~al.(2022{\natexlab{b}})Li, Liang, Zhao, Cui, Ouyang, Shao, Yu, and Yan]{li2021declip}
Yangguang Li, Feng Liang, Lichen Zhao, Yufeng Cui, Wanli Ouyang, Jing Shao, Fengwei Yu, and Junjie Yan.
\newblock Supervision exists everywhere: A data efficient contrastive language-image pre-training paradigm.
\newblock In \emph{ICLR}, 2022{\natexlab{b}}.

\bibitem[Li et~al.(2023{\natexlab{b}})Li, Fan, Hu, Feichtenhofer, and He]{li2023flip}
Yanghao Li, Haoqi Fan, Ronghang Hu, Christoph Feichtenhofer, and Kaiming He.
\newblock Scaling language-image pre-training via masking.
\newblock In \emph{CVPR}, 2023{\natexlab{b}}.

\bibitem[Liu et~al.(2023{\natexlab{a}})Liu, Yan, and Abbeel]{liu2024lqae}
Hao Liu, Wilson Yan, and Pieter Abbeel.
\newblock Language quantized autoencoders: Towards unsupervised text-image alignment.
\newblock In \emph{NeurIPS}, 2023{\natexlab{a}}.

\bibitem[Liu et~al.(2023{\natexlab{b}})Liu, Wang, Shao, Luo, Qiao, Shou, Zhang, and You]{liu2023mllm-a}
Yanqing Liu, Kai Wang, Wenqi Shao, Ping Luo, Yu Qiao, Mike~Zheng Shou, Kaipeng Zhang, and Yang You.
\newblock Mllms-augmented visual-language representation learning.
\newblock \emph{arXiv preprint arXiv:2311.18765}, 2023{\natexlab{b}}.

\bibitem[Loshchilov(2017)]{loshchilov2017adamw}
I Loshchilov.
\newblock Decoupled weight decay regularization.
\newblock \emph{arXiv preprint arXiv:1711.05101}, 2017.

\bibitem[Mu et~al.(2022)Mu, Kirillov, Wagner, and Xie]{mu2022slip}
Norman Mu, Alexander Kirillov, David Wagner, and Saining Xie.
\newblock Slip: Self-supervision meets language-image pre-training.
\newblock In \emph{ECCV}, 2022.

\bibitem[Muennighoff et~al.(2024)Muennighoff, Su, Wang, Yang, Wei, Yu, Singh, and Kiela]{muennighoff2024grit}
Niklas Muennighoff, Hongjin Su, Liang Wang, Nan Yang, Furu Wei, Tao Yu, Amanpreet Singh, and Douwe Kiela.
\newblock Generative representational instruction tuning.
\newblock \emph{arXiv preprint arXiv:2402.09906}, 2024.

\bibitem[Onoe et~al.(2024)Onoe, Rane, Berger, Bitton, Cho, Garg, Ku, Parekh, Pont-Tuset, Tanzer, et~al.]{onoe2024docci}
Yasumasa Onoe, Sunayana Rane, Zachary Berger, Yonatan Bitton, Jaemin Cho, Roopal Garg, Alexander Ku, Zarana Parekh, Jordi Pont-Tuset, Garrett Tanzer, et~al.
\newblock Docci: Descriptions of connected and contrasting images.
\newblock In \emph{ECCV}, 2024.

\bibitem[Radford et~al.(2021)Radford, Kim, Hallacy, Ramesh, Goh, Agarwal, Sastry, Askell, Mishkin, Clark, et~al.]{radford2021clip}
Alec Radford, Jong~Wook Kim, Chris Hallacy, Aditya Ramesh, Gabriel Goh, Sandhini Agarwal, Girish Sastry, Amanda Askell, Pamela Mishkin, Jack Clark, et~al.
\newblock Learning transferable visual models from natural language supervision.
\newblock In \emph{ICML}, 2021.

\bibitem[Rombach et~al.(2022)Rombach, Blattmann, Lorenz, Esser, and Ommer]{rombach2022stablediffusion}
Robin Rombach, Andreas Blattmann, Dominik Lorenz, Patrick Esser, and Bj{\"o}rn Ommer.
\newblock High-resolution image synthesis with latent diffusion models.
\newblock In \emph{CVPR}, 2022.

\bibitem[Sharma et~al.(2018)Sharma, Ding, Goodman, and Soricut]{sharma2018cc}
Piyush Sharma, Nan Ding, Sebastian Goodman, and Radu Soricut.
\newblock Conceptual captions: A cleaned, hypernymed, image alt-text dataset for automatic image captioning.
\newblock In \emph{ACL}, 2018.

\bibitem[Springer et~al.(2024)Springer, Kotha, Fried, Neubig, and Raghunathan]{springer2024echo}
Jacob~Mitchell Springer, Suhas Kotha, Daniel Fried, Graham Neubig, and Aditi Raghunathan.
\newblock Repetition improves language model embeddings.
\newblock \emph{arXiv preprint arXiv:2402.15449}, 2024.

\bibitem[Sun et~al.(2023)Sun, Fang, Wu, Wang, and Cao]{sun2023eva-clip}
Quan Sun, Yuxin Fang, Ledell Wu, Xinlong Wang, and Yue Cao.
\newblock Eva-clip: Improved training techniques for clip at scale.
\newblock \emph{arXiv preprint arXiv:2303.15389}, 2023.

\bibitem[Thapliyal et~al.(2022)Thapliyal, Tuset, Chen, and Soricut]{thapliyal2022crossmodal-3600}
Ashish~V Thapliyal, Jordi~Pont Tuset, Xi Chen, and Radu Soricut.
\newblock Crossmodal-3600: A massively multilingual multimodal evaluation dataset.
\newblock In \emph{EMNLP}, 2022.

\bibitem[Thrush et~al.(2022)Thrush, Jiang, Bartolo, Singh, Williams, Kiela, and Ross]{thrush2022winoground}
Tristan Thrush, Ryan Jiang, Max Bartolo, Amanpreet Singh, Adina Williams, Douwe Kiela, and Candace Ross.
\newblock Winoground: Probing vision and language models for visio-linguistic compositionality.
\newblock In \emph{CVPR}, 2022.

\bibitem[Tian et~al.(2023)Tian, Fan, Isola, Chang, and Krishnan]{tian2023stablerep}
Yonglong Tian, Lijie Fan, Phillip Isola, Huiwen Chang, and Dilip Krishnan.
\newblock Stablerep: Synthetic images from text-to-image models make strong visual representation learners.
\newblock In \emph{NeurIPS}, 2023.

\bibitem[Touvron et~al.(2023)Touvron, Martin, Stone, Albert, Almahairi, Babaei, Bashlykov, Batra, Bhargava, Bhosale, et~al.]{touvron2023llama2}
Hugo Touvron, Louis Martin, Kevin Stone, Peter Albert, Amjad Almahairi, Yasmine Babaei, Nikolay Bashlykov, Soumya Batra, Prajjwal Bhargava, Shruti Bhosale, et~al.
\newblock Llama 2: Open foundation and fine-tuned chat models.
\newblock \emph{arXiv preprint arXiv:2307.09288}, 2023.

\bibitem[Tsimpoukelli et~al.(2021)Tsimpoukelli, Menick, Cabi, Eslami, Vinyals, and Hill]{tsimpoukelli2021frozen}
Maria Tsimpoukelli, Jacob~L Menick, Serkan Cabi, SM Eslami, Oriol Vinyals, and Felix Hill.
\newblock Multimodal few-shot learning with frozen language models.
\newblock In \emph{NeurIPS}, 2021.

\bibitem[Visheratin(2023)]{visheratin2023nllb-clip}
Alexander Visheratin.
\newblock Nllb-clip--train performant multilingual image retrieval model on a budget.
\newblock \emph{arXiv preprint arXiv:2309.01859}, 2023.

\bibitem[Wang et~al.(2023)Wang, Yang, Huang, Yang, Majumder, and Wei]{wang2023e5mistralinst}
Liang Wang, Nan Yang, Xiaolong Huang, Linjun Yang, Rangan Majumder, and Furu Wei.
\newblock Improving text embeddings with large language models.
\newblock \emph{arXiv preprint arXiv:2401.00368}, 2023.

\bibitem[Xu et~al.(2022)Xu, De~Mello, Liu, Byeon, Breuel, Kautz, and Wang]{xu2022groupvit}
Jiarui Xu, Shalini De~Mello, Sifei Liu, Wonmin Byeon, Thomas Breuel, Jan Kautz, and Xiaolong Wang.
\newblock Groupvit: Semantic segmentation emerges from text supervision.
\newblock In \emph{CVPR}, 2022.

\bibitem[Yang et~al.(2022)Yang, Pan, Lin, Men, Zhang, Zhou, and Zhou]{yang2022cn-clip}
An Yang, Junshu Pan, Junyang Lin, Rui Men, Yichang Zhang, Jingren Zhou, and Chang Zhou.
\newblock Chinese clip: Contrastive vision-language pretraining in chinese.
\newblock \emph{arXiv preprint arXiv:2211.01335}, 2022.

\bibitem[Yang et~al.(2023)Yang, Deng, An, Li, Feng, Guo, Yang, and Liu]{yang2023alip}
Kaicheng Yang, Jiankang Deng, Xiang An, Jiawei Li, Ziyong Feng, Jia Guo, Jing Yang, and Tongliang Liu.
\newblock Alip: Adaptive language-image pre-training with synthetic caption.
\newblock In \emph{ICCV}, 2023.

\bibitem[Yao et~al.(2021)Yao, Huang, Hou, Lu, Niu, Xu, Liang, Li, Jiang, and Xu]{yao2021filip}
Lewei Yao, Runhui Huang, Lu Hou, Guansong Lu, Minzhe Niu, Hang Xu, Xiaodan Liang, Zhenguo Li, Xin Jiang, and Chunjing Xu.
\newblock Filip: Fine-grained interactive language-image pre-training.
\newblock \emph{arXiv preprint arXiv:2111.07783}, 2021.

\bibitem[Yu et~al.(2022)Yu, Wang, Vasudevan, Yeung, Seyedhosseini, and Wu]{yu2022coca}
Jiahui Yu, Zirui Wang, Vijay Vasudevan, Legg Yeung, Mojtaba Seyedhosseini, and Yonghui Wu.
\newblock Coca: Contrastive captioners are image-text foundation models.
\newblock \emph{arXiv preprint arXiv:2205.01917}, 2022.

\bibitem[Yu et~al.(2023)Yu, Cheng, Wang, Kumar, Macherey, Huang, Ross, Essa, Bisk, Yang, et~al.]{yu2024spae}
Lijun Yu, Yong Cheng, Zhiruo Wang, Vivek Kumar, Wolfgang Macherey, Yanping Huang, David Ross, Irfan Essa, Yonatan Bisk, Ming-Hsuan Yang, et~al.
\newblock Spae: Semantic pyramid autoencoder for multimodal generation with frozen llms.
\newblock In \emph{NeurIPS}, 2023.

\bibitem[Zhai et~al.(2022)Zhai, Wang, Mustafa, Steiner, Keysers, Kolesnikov, and Beyer]{zhai2022lit}
Xiaohua Zhai, Xiao Wang, Basil Mustafa, Andreas Steiner, Daniel Keysers, Alexander Kolesnikov, and Lucas Beyer.
\newblock Lit: Zero-shot transfer with locked-image text tuning.
\newblock In \emph{CVPR}, 2022.

\bibitem[Zhai et~al.(2023)Zhai, Mustafa, Kolesnikov, and Beyer]{zhai2023siglip}
Xiaohua Zhai, Basil Mustafa, Alexander Kolesnikov, and Lucas Beyer.
\newblock Sigmoid loss for language image pre-training.
\newblock In \emph{ICCV}, 2023.

\bibitem[Zhang et~al.(2024{\natexlab{a}})Zhang, Chang, and Li]{zhang2024pretcot}
Bowen Zhang, Kehua Chang, and Chunping Li.
\newblock Simple techniques for enhancing sentence embeddings in generative language models.
\newblock In \emph{ICIC}, 2024{\natexlab{a}}.

\bibitem[Zhang et~al.(2024{\natexlab{b}})Zhang, Zhang, Dong, Zang, and Wang]{zhang2024long-clip}
Beichen Zhang, Pan Zhang, Xiaoyi Dong, Yuhang Zang, and Jiaqi Wang.
\newblock Long-clip: Unlocking the long-text capability of clip.
\newblock In \emph{ECCV}, 2024{\natexlab{b}}.

\bibitem[Zheng et~al.(2024)Zheng, Zhang, Wu, Lu, Ma, Jin, Chen, and Shen]{zheng2024dreamlip}
Kecheng Zheng, Yifei Zhang, Wei Wu, Fan Lu, Shuailei Ma, Xin Jin, Wei Chen, and Yujun Shen.
\newblock Dreamlip: Language-image pre-training with long captions.
\newblock In \emph{ECCV}, 2024.

\bibitem[Zhu et~al.(2024)Zhu, Wei, and Lu]{zhu2024v2l-tokenizer}
Lei Zhu, Fangyun Wei, and Yanye Lu.
\newblock Beyond text: Frozen large language models in visual signal comprehension.
\newblock In \emph{CVPR}, 2024.

\bibitem[Zhuang et~al.(2024)Zhuang, Ma, Koopman, Lin, and Zuccon]{zhuang2024promptreps}
Shengyao Zhuang, Xueguang Ma, Bevan Koopman, Jimmy Lin, and Guido Zuccon.
\newblock Promptreps: Prompting large language models to generate dense and sparse representations for zero-shot document retrieval.
\newblock In \emph{EMNLP}, 2024.

\end{thebibliography}
}

\clearpage
\clearpage
\setcounter{page}{1}
\maketitlesupplementary

\setcounter{section}{0}
\setcounter{figure}{0}
\setcounter{table}{0}
\renewcommand\thesection{\Alph{section}}
\renewcommand\thefigure{\Alph{section}\arabic{figure}}
\renewcommand\thetable{\Alph{section}\arabic{table}}

\noindent This document is structured as follows:
\begin{itemize}[topsep=1pt, partopsep=9pt, itemsep=-1pt, parsep=0.5ex]
    \item In Section~\ref{sec:hyper}, we provide details of the hyper-parameters.
    \item In Section~\ref{sec:prompts}, we present the full list of multifaceted prompts used in our experiments.
    \item In Section~\ref{sec:additional}, we conduct additional experiments, including: 1) multilingual image-to-text retrieval, 2) vision-language compositionality, 3) few-shot image comprehension, 4) ablations on individual semantic levels, and 5) preliminary exploration of scalability.
    \item In Section~\ref{sec:visualizations}, we provide visualizations of text-to-image retrieval.
\end{itemize}

\section{Hyper-Parameters}
\label{sec:hyper}
Table~\ref{tab:hyper_cc3m} and \ref{tab:hyper_yfcc15m} provide the pre-training hyperparameters used for CC3M and YFCC15M, respectively. These hyper-parameters are set similarly to those used in previous methods~\cite{fan2024laclip, zheng2024dreamlip} for fair comparisons.
\begin{table}[h]
\small
\centering
\begin{NiceTabular}{l|l}
Config & Value \\
\midrule[1pt]
Batch size & 4,096 \\
Optimizer & AdamW~\cite{loshchilov2017adamw} \\
Learning rate & 5e-4 \\
Weight decay & 0.5 \\
Adam $\beta_1$, $\beta_2$ & 0.9, 0.98 \\
Adam $\epsilon$ & 1e-8 \\
Total epochs & 32 \\
Warm up iterations & 2,000 \\
Learning rate schedule & cosine decay \\
\end{NiceTabular}
\vspace{-5pt}
\caption{\textbf{Hyper-parameters for CC3M.}}
\label{tab:hyper_cc3m}
\vspace{-.5em}
\end{table}
\begin{table}[h]
\small
\centering
\begin{NiceTabular}{l|l}
Config & Value \\
\midrule[1pt]
Batch size & 8,192 \\
Optimizer & AdamW~\cite{loshchilov2017adamw} \\
Learning rate & 5e-4 \\
Weight decay & 0.2 \\
Adam $\beta_1$, $\beta_2$ & 0.9, 0.98 \\
Adam $\epsilon$ & 1e-8 \\
Total epochs & 32 \\
Warm up iterations & 2,000 \\
Learning rate schedule & cosine decay \\
\end{NiceTabular}
\vspace{-5pt}
\caption{\textbf{Hyper-parameters for YFCC15M.}}
\label{tab:hyper_yfcc15m}
\vspace{-.5em}
\end{table}

\section{Full List of Multifaceted Prompts}
\label{sec:prompts}
We provide all the designed multifaceted prompts here. The distinct part in each prompt is \textcolor{blue1}{marked}. We also annotate the \textbf{default} prompts for long and short text input.

\paragraph{Entity Level:}
\begin{itemize}[noitemsep, leftmargin=*]
    \item Detailed image description: ``$y_i$". After thinking step by step, \textcolor{blue1}{the category of the main object in this image} means in just one word:" \textbf{(default for long input)}
    \item Detailed image description: ``$y_i$". After thinking step by step, \textcolor{blue1}{the prominent characteristic or pattern of the main object in this image} means in just one word:" \textbf{(default for long input)}
    \item Detailed image description: ``$y_i$". After thinking step by step, \textcolor{blue1}{the category of the minor object in this image} means in just one word:" \textbf{(default for long input)}
    \item Detailed image description: ``$y_i$". After thinking step by step, \textcolor{blue1}{the prominent characteristic or pattern of the minor object in this image} means in just one word:" \textbf{(default for long input)}
\end{itemize}

\paragraph{Interaction Level:}
\begin{itemize}[noitemsep, leftmargin=*]
    \item Detailed image description: ``$y_i$". After thinking step by step, \textcolor{blue1}{the primary action or event taking place in this image} means in just one word:" \textbf{(default for long input)}
    \item Detailed image description: ``$y_i$". After thinking step by step, \textcolor{blue1}{the positioning layout or spatial relationship in this image} means in just one word:"
\end{itemize}

\paragraph{Scene Level:}
\begin{itemize}[noitemsep, leftmargin=*]
    \item Detailed image description: ``$y_i$". After thinking step by step, \textcolor{blue1}{this image description} means in just one word:" \textbf{(default for long and short inputs)}
    \item Detailed image description: ``$y_i$". After thinking step by step, \textcolor{blue1}{the overall atmosphere or emotion of this image} means in just one word:" \textbf{(default for long input)}
    \item Detailed image description: ``$y_i$". After thinking step by step, \textcolor{blue1}{the dominant color or color combination of this image} means in just one word:"
\end{itemize}

\section{Additional Experiments}
\label{sec:additional}

\begin{figure}[h]
    \includegraphics[width=0.99\linewidth]{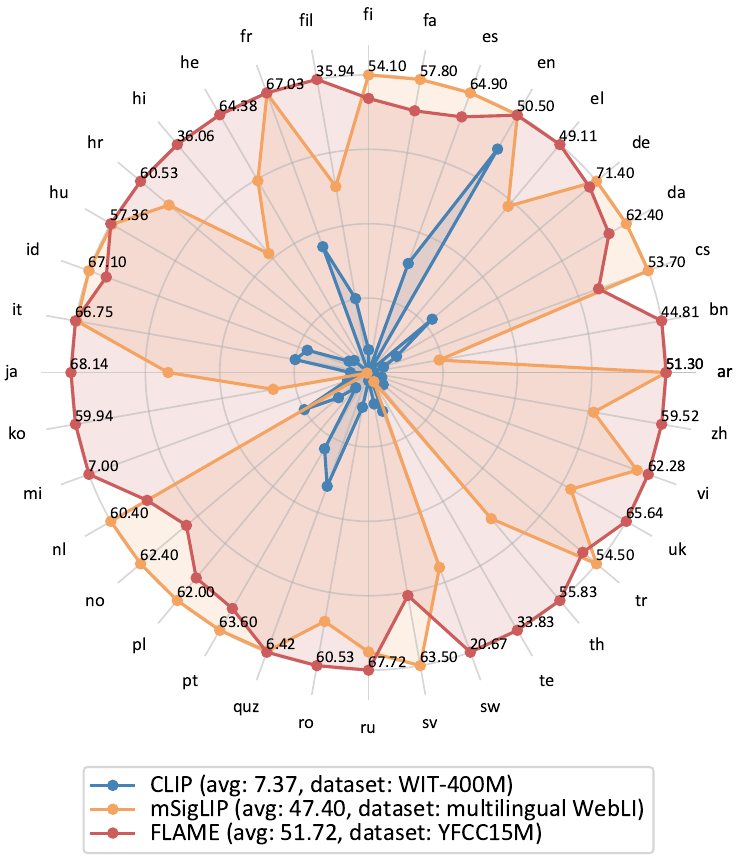}
\vspace{-5pt}
    \caption{\textbf{Multilingual zero-shot image-to-text retrieval recall@1 results on Crossmodal-3600.}}
    \label{fig:multilingual_supp}
\vspace{-1em}
\end{figure}
\begin{table*}[t]
\centering
\resizebox{\linewidth}{!}{
\begin{NiceTabular}{l|l|ccc|cc|ccc|cc|c}
& & \multicolumn{3}{c}{Winoground}& \multicolumn{2}{c}{SugarCrepe-Add} & \multicolumn{3}{c}{SugarCrepe-Replace}&  \multicolumn{2}{c}{SugarCrepe-Swap} & \\
Method & Dataset & Image & Text & Both & Attribute & Object & Attribute & Object & Relation & Attribute & Object & Avg. \\
\midrule[1pt]
CLIP~\cite{radford2021clip} & WIT-400M & 10.8 & 25.0 & 7.5 & 66.8 & 78.5 & 81.1 & 93.4 & 66.9 & 64.6 & 60.0 & 55.5 \\
DreamLIP~\cite{zheng2024dreamlip} & YFCC15M & 14.7 & 26.2 & 9.7 & 78.3 & 80.3 & 81.3 & 91.0 & 72.9 & \textbf{77.5} & \textbf{66.9} & 59.9 \\
FLAME & YFCC15M & \cellcolor{gray!10}\textbf{18.3} & \cellcolor{gray!10}\textbf{34.5} & \cellcolor{gray!10}\textbf{13.2} & \cellcolor{gray!10}\textbf{82.4} & \cellcolor{gray!10}\textbf{87.8} & \cellcolor{gray!10}\textbf{85.8} & \cellcolor{gray!10}\textbf{94.1} & \cellcolor{gray!10}\textbf{79.9} & \cellcolor{gray!10}67.6 & \cellcolor{gray!10}66.1 & \cellcolor{gray!10}\textbf{63.0} \\

\end{NiceTabular}
}
\vspace{-5pt}
\caption{\textbf{Vision-language compositionality.} FLAME demonstrates better fine-grained scene understanding capability.}
\label{tab:compositionality}
\vspace{-.5em}
\end{table*}
\begin{table*}[t]
\small
\centering
\begin{NiceTabular}{l|l|ccccc|ccccc}
\multirow{2}{*}{Method} & N-Way K-shot & 2-1 & 2-3 & 2-1 & 2-1 & \multirow{2}{*}{Avg.} & 5-1 & 5-3 & 5-1 & 5-1 & \multirow{2}{*}{Avg.} \\
& \# Repetitions & 0 & 0 & 1 & 3 & & 0 & 0 & 1 & 3 & \\
\midrule[1pt]
Frozen~\cite{tsimpoukelli2021frozen} & - & 33.7 & 66.0 & 63.0 & 65.0 & 56.9 & 14.5 & 34.7 & 33.8 & 33.3 & 29.1 \\
LQAE~\cite{liu2024lqae} & GPT-3.5 & 35.2 & 68.2 & 68.5 & 68.7 & 60.2 & 15.7 & 35.9 & 31.9 & 36.4 & 30.0 \\
V2L-Tokenizer~\cite{zhu2024v2l-tokenizer} & Llama-2-7B & 76.3 & 91.2 & 84.0 & 84.4 & 84.0 & 44.8 & \textbf{91.8} & \textbf{73.9} & \textbf{82.2} & \textbf{73.2} \\
SPAE~\cite{yu2024spae} & PaLM-2-340B & \textbf{84.8} & \textbf{92.5} & \underline{84.8} & \underline{85.2} & \textbf{86.8} & \textbf{65.1} & 73.7 & \underline{66.4} & 67.0 & 68.1 \\
FLAME & Mistral-7B & \cellcolor{gray!10}\underline{83.3} & \cellcolor{gray!10}\underline{91.7} & \cellcolor{gray!10}\textbf{85.7} & \cellcolor{gray!10}\textbf{86.3} & \cellcolor{gray!10}\textbf{86.8} & \cellcolor{gray!10}\underline{55.7} & \cellcolor{gray!10}\underline{82.1} & \cellcolor{gray!10}65.7 &\cellcolor{gray!10}\underline{70.1} & \cellcolor{gray!10}\underline{68.4} \\
\end{NiceTabular}
\vspace{-5pt}
\caption{\textbf{Few-shot image comprehension on MiniImageNet.} FLAME's performance is either comparable to or surpasses that of SPAE.}
\label{tab:fs_cls}
\vspace{-.5em}
\end{table*}


\subsection{Multilingual Image-to-Text Retrieval}
We provide image-to-text retrieval results on Crossmodal-3600~\cite{thapliyal2022crossmodal-3600} in Figure~\ref{fig:multilingual_supp}.

\subsection{Vision-Language Compositionality} 

To evaluate the vision-language compositional understanding (i.e. whether the model understands the fine-grained atomic concepts that compose the scene), we conduct experiments on the Winoground~\cite{thrush2022winoground} and SugarCrepe~\cite{hsieh2024sugarcrepe} benchmarks, with results presented in Table~\ref{tab:compositionality}. 
As shown, FLAME trained on YFCC15M significantly outperforms WIT-400M-trained CLIP~\cite{radford2021clip} across all tasks, especially in relation understanding. Furthermore, when compared to the recent work~\cite{zheng2024dreamlip} that utilizes multiple short captions, FLAME retains an advantage on 8 out of 10 tasks, achieving an average improvement of 3.1\%. These results reveal that our multifaceted training promotes fine-grained semantic learning.

\begin{table*}[t]
\small
\centering
\begin{NiceTabular}{c|c|cccccccccc|c|c}
\rotatebox[origin=c]{0}{Backbone} & \rotatebox[origin=c]{0}{Method} & \rotatebox[origin=l]{90}{Food-101} & \rotatebox[origin=l]{90}{CIFAR-10} & \rotatebox[origin=l]{90}{CIFAR-100} &	\rotatebox[origin=l]{90}{SUN397} &	\rotatebox[origin=l]{90}{Cars} & \rotatebox[origin=l]{90}{Aircraft} & \rotatebox[origin=l]{90}{DTD} & \rotatebox[origin=l]{90}{Pets} & \rotatebox[origin=l]{90}{Caltech-101} & \rotatebox[origin=l]{90}{Flowers} & \rotatebox[origin=l]{90}{Average} & \rotatebox[origin=l]{90}{ImageNet} \\
\midrule[1pt]
\multirow{2}{*}{ViT-B/16} & CLIP~\cite{radford2021clip} & 35.0 & 67.1 & 34.8 & 42.0 & 5.1 & 6.3 & 13.9 & 20.4 & 54.5 & 44.3 & 32.3 & 34.1 \\
& FLAME & \cellcolor{gray!10}\textbf{61.8} & \cellcolor{gray!10}\textbf{86.1} & \cellcolor{gray!10}\textbf{56.7} & \cellcolor{gray!10}\textbf{66.8} & \cellcolor{gray!10}\textbf{10.7} & \cellcolor{gray!10}\textbf{10.3} & \cellcolor{gray!10}\textbf{54.9} & \cellcolor{gray!10}\textbf{40.7} & \cellcolor{gray!10}\textbf{78.9} & \cellcolor{gray!10}\textbf{51.7} & \cellcolor{gray!10}\textbf{51.9} & \cellcolor{gray!10}\textbf{51.5} \\
\midrule
\multirow{2}{*}{ViT-L/14} & CLIP~\cite{radford2021clip} & 42.2 & 66.8 & 33.1 & 45.3 & 2.7 & 2.3 & 19.7 & 26.4 & 65.1 & 53.1 & 35.7 & 36.3 \\
& FLAME & \cellcolor{gray!10}\textbf{68.3} & \cellcolor{gray!10}\textbf{87.5} & \cellcolor{gray!10}\textbf{58.4} & \cellcolor{gray!10}\textbf{68.3} & \cellcolor{gray!10}\textbf{14.6} & \cellcolor{gray!10}\textbf{11.3} & \cellcolor{gray!10}\textbf{56.3} & \cellcolor{gray!10}\textbf{43.8} & \cellcolor{gray!10}\textbf{80.3} & \cellcolor{gray!10}\textbf{56.3} & \cellcolor{gray!10}\textbf{54.5} & \cellcolor{gray!10}\textbf{54.8} \\
\end{NiceTabular}

\vspace{-5pt}
\caption{\textbf{Zero-shot classification on YFCC15M with ViT-L/14.} FLAME consistently demonstrates substantial advantages over CLIP.}
\label{tab:scale_zs_cls}
\vspace{-.5em}
\end{table*}
\begin{table}[t]
\small
\centering
\resizebox{\linewidth}{!}{
\begin{NiceTabular}{c|c|ccccccc|c}
\rotatebox[origin=c]{0}{Backbone} & \rotatebox[origin=c]{0}{Method} & \rotatebox[origin=l]{90}{Food-101} & \rotatebox[origin=l]{90}{CIFAR-10} & \rotatebox[origin=l]{90}{CIFAR-100} &	\rotatebox[origin=l]{90}{Cars} & \rotatebox[origin=l]{90}{Aircraft} & \rotatebox[origin=l]{90}{DTD} & \rotatebox[origin=l]{90}{Caltech-101} & \rotatebox[origin=l]{90}{Average} \\
\midrule[1pt]
\multirow{2}{*}{ViT-B/16} & CLIP~\cite{radford2021clip} & 77.2 & 88.5 & 66.4 & 29.0 & 25.5 & 65.2 & 82.4 & 62.0 \\
& FLAME & \cellcolor{gray!10}\textbf{85.9} & \cellcolor{gray!10}\textbf{95.0} & \cellcolor{gray!10}\textbf{81.0} & \cellcolor{gray!10}\textbf{54.3} & \cellcolor{gray!10}\textbf{39.3} & \cellcolor{gray!10}\textbf{76.8} & \cellcolor{gray!10}\textbf{92.5} & \cellcolor{gray!10}\textbf{75.0} \\
\midrule
\multirow{2}{*}{ViT-L/14} & CLIP~\cite{radford2021clip} & 74.4 & 88.9 & 69.7 & 27.3 & 26.1 & 63.7 & 86.4 & 62.3 \\
& FLAME & \cellcolor{gray!10}\textbf{88.0} & \cellcolor{gray!10}\textbf{95.9} & \cellcolor{gray!10}\textbf{81.7} & \cellcolor{gray!10}\textbf{64.3} & \cellcolor{gray!10}\textbf{46.3} & \cellcolor{gray!10}\textbf{78.4} & \cellcolor{gray!10}\textbf{93.8} & \cellcolor{gray!10}\textbf{78.3} \\
\end{NiceTabular}
}
\vspace{-5pt}
\caption{\textbf{Linear-probe classification on YFCC15M with ViT-L/14.} FLAME benefits from the increased scale of the visual backbone.}
\label{tab:scale_lp_cls}
\vspace{-.5em}
\end{table}

\subsection{Few-Shot Image Comprehension}

Following SPAE~\cite{yu2024spae} and V2L-Tokenizer~\cite{zhu2024v2l-tokenizer}, we conduct few-shot image comprehension experiments on both 2-way and 5-way MiniImageNet benchmarks. Our evaluation methodology involves converting image patches into words and then using the original LLM for in-context reasoning, without additional fine-tuning. Specifically, given a few-shot example image set $\{x_i\}_{i=1}^S$, we perform the vocabulary mapping process to obtain $\{C_i\}_{i=1}^S$, as mentioned in Section~\ref{subsec:semantic_interpretability}. By pairing each $C_i$ with its corresponding text answer $A_i$, we construct the in-context example $E = \{\langle C_i, A_i \rangle\}_{i=1}^S$. We input this example and the new context $\Tilde{C}$ to the LLM, which then outputs the answer $\Tilde{A}$ for verification of correctness. This reasoning process can be formulated as $\Tilde{A} = \mathrm{LLM}(E, \Tilde{C})$. As shown in Table~\ref{tab:fs_cls}, FLAME achieves average accuracies of 86.8\% and 68.4\% on 2-way and 5-way scenarios, respectively. These results are comparable to or surpass SPAE~\cite{yu2024spae}. FLAME achieves these results using a smaller 7B backbone than SPAE's 340B model.

\subsection{Ablations on Individual Semantic Levels}
\setlength{\columnsep}{5pt}
\begin{wraptable}[7]{r}{0.55\linewidth}
\vspace{-5mm}
\small
\centering
\resizebox{\linewidth}{!}{
\begin{NiceTabular}{l|cc}
 & \multicolumn{2}{c}{Long Avg.} \\
Level & I2T & T2I \\
\midrule[1pt]
+Scene & 50.5 & 47.4 \\
+Interaction & 52.9 & 50.3 \\
+Entity & 64.1 & 62.0 \\
\end{NiceTabular}
}
\vspace{-8pt}
\caption{\footnotesize\textbf{Semantic decomposition.}}
\label{tab:semantic_decompositioin}
\vspace{-1em}
\end{wraptable}
To further clarify the reasonability of our semantic decomposition, we conduct ablations by orderly increasing individual semantic levels. The results are presented in Table~\ref{tab:semantic_decompositioin}. Evidently, a systematic increase in semantic levels consistently yields positive outcomes. This also indicates that our rational prompt design is generalizable and does not lead to detrimental caption-prompt misalignment.

\subsection{Preliminary Exploration of Scalability} 
To explore the scalability of FLAME, we increase the size of the visual encoder from the default ViT-B/16 to ViT-L/14 and train it on YFCC15M.

Table~\ref{tab:scale_zs_cls} presents the zero-shot classification results, where FLAME continues to exhibit a substantial advantage over CLIP when using ViT-L/14, as evidenced by an improvement in ImageNet top-1 accuracy of 18.5\% and an increase in downstream average accuracy of 18.8\%. We also perform linear-probe classification experiments, with results shown in Table~\ref{tab:scale_lp_cls}.
\begin{table}[h]
\small
\centering
\resizebox{\linewidth}{!}{
\begin{NiceTabular}{l|l|cccccc}
& & \multicolumn{6}{c}{Evaluation} \\
Method & Training & ar & en & it & jp & zh & avg \\
\midrule[1pt]
CLIP~\cite{radford2021clip} & en & 0.6 & \textbf{75.5} & 27.6 & 4.5 & 1.9 & 22.0 \\
CN-CLIP~\cite{yang2022cn-clip} & en+zh & 0.1 & 32.5 & 8.0 & 16.3 & \textbf{53.4} & 22.1 \\
NLLB-CLIP~\cite{visheratin2023nllb-clip} & en+multi & 25.8 & 36.7 & 27.4 & 23.9 & 24.3 & 27.6 \\
FLAME & en & \cellcolor{gray!10}\textbf{34.9} & \cellcolor{gray!10}54.8 & \cellcolor{gray!10}\textbf{44.9} & \cellcolor{gray!10}\textbf{39.6} & \cellcolor{gray!10}42.6 & \cellcolor{gray!10}\textbf{43.4} \\

\end{NiceTabular}
}
\vspace{-5pt}
\caption{\textbf{Multilingual ImageNet1k classification with the ``large" model variant.} CLIP is trained on WIT-400M. CN-CLIP is pre-trained on WIT-400M and fine-tuned on 200M Chinese data. NLLB-CLIP is pre-trained on LAION-2B and fine-tuned on LAION-COCO-NLLB with 200 languages. FLAME is trained on YFCC15M.}
\label{tab:scale_multilingual_cls}
\vspace{-.5em}
\end{table}
In multilingual scenarios, the use of this larger backbone also proves beneficial, yielding a 2.8\% improvement in average accuracy on the multilingual ImageNet1k classification benchmark. Detailed results and comparisons are provided in Table~\ref{tab:scale_multilingual_cls}.

These notable performance enhancements illustrate the scalability of FLAME, paving the way for future large-scale language-image pre-training.

\section{Visualizations} 
\label{sec:visualizations}

\begin{figure*}[t]
\centering
\includegraphics[width=\linewidth]{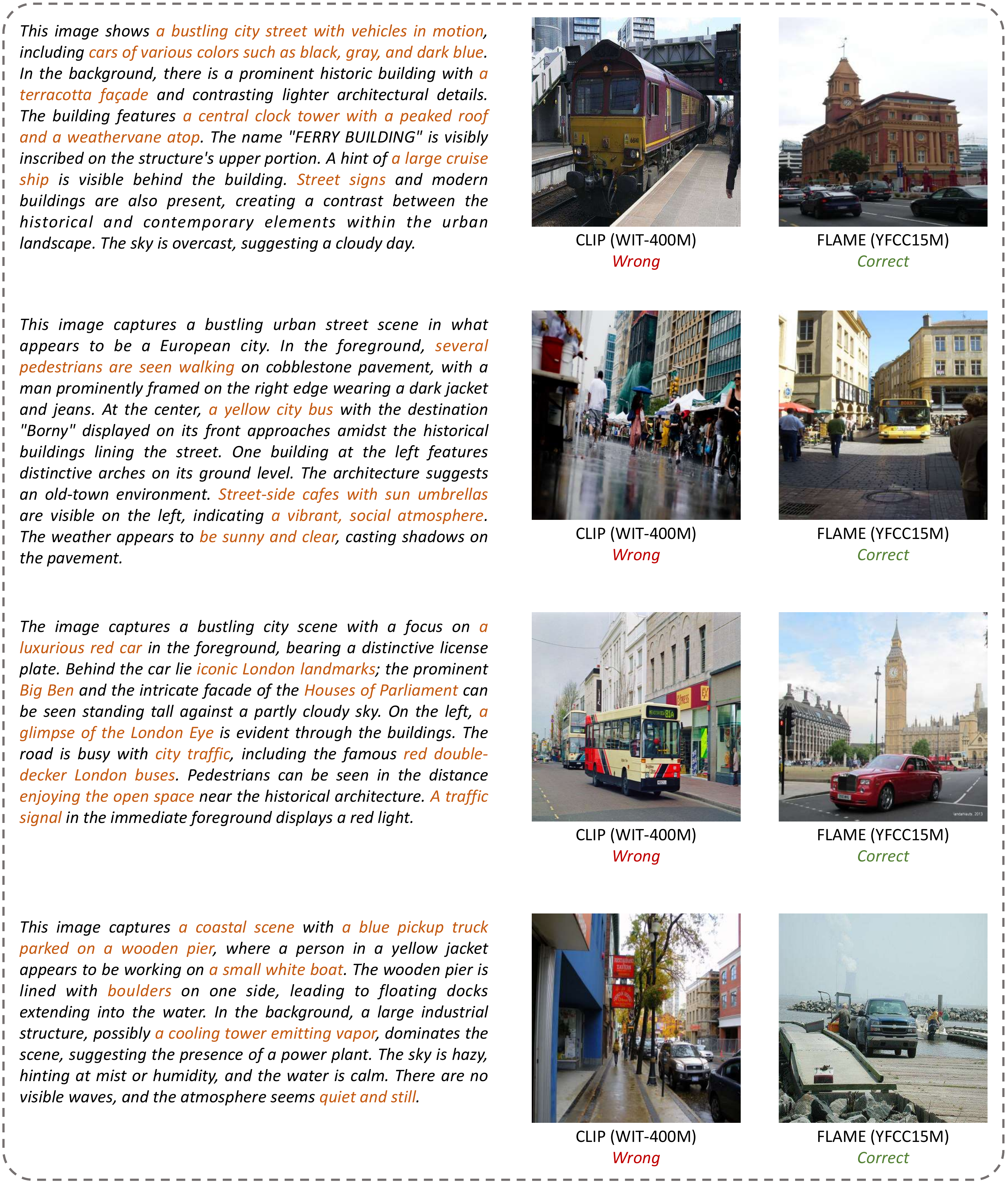}
    \caption{\textbf{Visualizations of text-to-image retrieval on Urban-1k.}}
    \label{fig:visualizations}
\end{figure*}

Figure~\ref{fig:visualizations} presents some visualizations of text-to-image retrieval on Urban-1k, comparing the results of FLAME with those of CLIP.

\end{document}